\pdfoutput=1
\documentclass[11pt]{article}
\usepackage[final]{acl} % review, final, preprint

\usepackage{times}
\usepackage{latexsym}
\usepackage[T1]{fontenc}
\usepackage[utf8]{inputenc}
\usepackage{microtype}
\usepackage{inconsolata}
\usepackage{graphicx}
\usepackage{amsmath}
\usepackage{booktabs}
\usepackage{multirow}
\usepackage{subcaption}
\usepackage{enumitem}

\title{SFTMix: Elevating Language Model Instruction Tuning with Mixup Recipe}

\author{Yuxin Xiao$^{1}$\thanks{Work done during an internship at Zoom.}, Shujian Zhang$^{2}$, Wenxuan Zhou$^{2}$, Marzyeh Ghassemi$^{1}$, Sanqiang Zhao$^{2}$\thanks{Corresponding author.}\\
$^{1}$Massachusetts Institute of Technology, $^{2}$Zoom Video Communications\\
\texttt{yuxin102@mit.edu}}

\begin{document}
\maketitle
\begin{abstract}
To acquire instruction-following capabilities, large language models (LLMs) undergo instruction tuning, where they are trained on instruction-response pairs using next-token prediction (NTP).
Efforts to improve instruction tuning often focus on higher-quality supervised fine-tuning (SFT) datasets, typically requiring data filtering with proprietary LLMs or human annotation.
In this paper, we take a different approach by proposing SFTMix, a novel Mixup-based recipe that elevates LLM instruction tuning without relying on well-curated datasets. 
We observe that LLMs exhibit uneven confidence across the semantic representation space.
We argue that examples with different confidence levels should play distinct roles in instruction tuning: Confident data is prone to overfitting, while unconfident data is harder to generalize.
Based on this insight, SFTMix leverages training dynamics to identify examples with varying confidence levels.
We then interpolate them to bridge the confidence gap and apply a Mixup-based regularization to support learning on these additional, interpolated examples.
We demonstrate the effectiveness of SFTMix in both instruction-following and healthcare-specific SFT tasks, with consistent improvements across LLM families and SFT datasets of varying sizes and qualities.
Extensive analyses across six directions highlight SFTMix's compatibility with data selection, adaptability to compute-constrained scenarios, and scalability to broader applications.
\end{abstract}
\section{Introduction} \label{sec:1}

LLMs have recently achieved strong performance across diverse natural language processing (NLP) tasks~\citep{zhao2023survey, minaee2024large}. 
After being pre-trained on large corpora of raw text, LLMs undergo a critical instruction-tuning stage~\citep{ouyang2022training, zhang2023instruction} to develop their instruction-following capabilities based on SFT datasets~\citep{alpaca, wang2023self, xu2024wizardlm}.
During this stage, LLMs are usually trained via NTP, where they predict the next token in a response given both the instruction and the preceding tokens in that response.

Previous research in this field has predominantly focused on enhancing the quality of instruction-tuning datasets. 
One line of research direction seeks to better understand the intrinsic properties of these datasets~\citep{kung2023active, lin2024not} and selects informative instruction-response pairs through heuristic-based filters~\citep{zhao2024long} or LLM scoring~\citep{chen2024alpagasus}.
Another line of work generates high-quality responses by querying advanced proprietary LLMs~\citep{chen2024alpagasus} or relying on human annotators~\citep{zhou2023lima}.

In this paper, we take a different approach by proposing SFTMix, a novel Mixup-based~\citep{zhang2018mixup} recipe to elevate LLM instruction tuning without the need for well-curated datasets.
Our design is motivated by the observation that an LLM's confidence distribution is uneven across the semantic representation space.
Since confident data is prone to overfitting~\citep{zhang2021mixup, han2024selective} and unconfident data is harder to generalize~\citep{elsayed2018large, jiang2018predicting}, we argue that data with varying confidence levels should play distinct roles in instruction tuning.
Hence, we first derive an LLM's confidence from its training dynamics~\citep{swayamdipta2020dataset} and divide the SFT dataset into confident and unconfident subsets accordingly.
We then linearly interpolate between these subsets and introduce a Mixup-based regularization to support learning on these additional, interpolated examples.
By propagating supervision signals across confidence regions~\citep{bengio2009curriculum, chapelle2009semi, sohn2020fixmatch} and encouraging linear behavior between them~\citep{zhang2018mixup, verma2019manifold}, our recipe mitigates overfitting in confident examples while enhancing generalization in unconfident ones during LLM instruction tuning.

We demonstrate the effectiveness of our proposed SFTMix recipe in both instruction-following and domain-specific SFT settings.
In particular, SFTMix significantly outperforms the conventional NTP baseline in both MT-Bench~\citep{zheng2023judging} and AlpacaEval-2~\citep{dubois2024lengthcontrolled}, with consistent improvements across LLM families (i.e., Llama~\citep{dubey2024llama}, Mistral~\citep{jiang2023mistral}, and Qwen~\citep{hui2024qwen25}) and SFT datasets of varying sizes and qualities (i.e., Alpaca-52K~\citep{alpaca}, UltraChat-200K~\citep{tunstall2023zephyr}, and Tulu3-939K~\citep{lambert2024tulu}).
Moreover, in the healthcare domain, Llama-3.1-8B and Mistral-7B-v0.1, instruction-tuned on MedAlpaca-263K~\citep{han2023medalpaca} using SFTMix, increase the accuracy by an average of $1.5\%$ absolutely across four question-answering benchmarks~\citep{jin2019pubmedqa, jin2021disease, pal2022medmcqa}.

In addition, we conduct in-depth analyses across six directions to highlight SFTMix’s versatility and scalability in LLM instruction tuning.
Our results validate the importance of confidence-based data splitting for effective Mixup and show that Mixup works best as a regularization alongside NTP.
Moreover, we demonstrate that SFTMix integrates seamlessly with data selection~\citep{chen2024alpagasus, zhao2024long}, adapts well to compute-constrained scenarios~\citep{hu2022lora}, and scales effectively to broader applications~\citep{burns2024weaktostrong}.

We summarize our contributions as follows:
\begin{itemize}[noitemsep,topsep=0pt]
    \item We introduce SFTMix, a novel recipe to elevate LLM instruction tuning without relying on well-curated SFT datasets, by interpolating semantic regions of varying confidence levels and applying a Mixup-based regularization.
    \item We show that SFTMix outperforms the NTP baseline across various instruction-following and healthcare-specific SFT tasks, with consistent improvements across LLM families and SFT datasets of varying sizes and qualities.
    \item Extensive analyses across six directions highlight that SFTMix is compatible with data selection, adaptable to compute-constrained scenarios, and scalable to broader applications.
\end{itemize}
% \vspace{-0.5mm}
\section{Related Work} \label{sec:2}
% \vspace{-0.25mm}

\paragraph{LLM Instruction Tuning.}
To align LLMs with user intents or domain-specific tasks,~\citet{ouyang2022training} propose instruction-tuning LLMs on human-annotated demonstrations using supervised learning. 
The conventional NTP paradigm trains LLMs to predict response tokens sequentially given instruction-response pairs~\citep{zhang2023instruction}.
Enhancements include adding noise to token embeddings~\citep{jainneftune}, commonality-aware partition~\citep{rao2024commonit}, and explicitly modeling instructions~\citep{shi2024instruction}.
Previous work~\citep{vicuna2023, ding2023enhancing, alpaca, wang2023self, xu2024wizardlm} collects instruction-following datasets via LLM distillation or crowd-sourced user conversations.
To improve data quality, researchers employ heuristic-based filters~\citep{schoch2023data, zhao2024long, yang2025measuring}, importance weighting~\citep{xie2023data, xia2024less, chen2024take}, LLM scoring~\citep{chen2024alpagasus}, and human curation~\citep{zhou2023lima}.
Other studies explore the intrinsic properties of SFT datasets~\citep{kung2023active, lin2024not, fu2025t, zhang2025best}, gradient-based methods~\citep{wang2025nice, zhao2025beyond}, and reinforcement learning from human feedback~\citep{rafailov2023direct, ethayarajh2024model, zeng2024token}.
However, acquiring high-quality SFT data often entails substantial computational and labor costs. 
This paper aims to optimize data utilization through confidence-aware data interpretation and elevate instruction tuning beyond NTP without relying on well-curated datasets.

\paragraph{Data Characterization via Training Dynamics.}
Data characterization~\citep{albalak2024survey, wang2024survey} analyzes training data quality to improve downstream model performance. 
In particular,~\citet{swayamdipta2020dataset} leverage training dynamics from a pre-trained language model~\citep{liu2019roberta} to create data maps.
This idea has inspired advances in active learning~\citep{zhang2021cartography, zhang2022allsh, kung2023active}, curriculum learning~\citep{christopoulou2022training, lin2024not, poesina-etal-2024-novel}, dataset pruning~\citep{chimoto-etal-2024-critical, he2024large, lin2024not, seedat2024curated}, and LLM scaling~\citep{mircea2025training, qi2025evolm, zhang2025training}.
Here, we apply training dynamics to causal language generation by categorizing an SFT dataset into confident and unconfident subsets, which facilitates the subsequent Mixup-based regularization during LLM instruction tuning.

\paragraph{Mixup-Based Learning.}
To mitigate memorization and adversarial sensitivity, Mixup~\citet{zhang2018mixup} trains models on convex combinations of paired inputs and labels.
Its variants~\citep{verma2019manifold, hendrycks2020augmix, uddin2021saliencymix, choi2022tokenmixup} interpolate feature representations at various stages, guided by different training signals.
Theoretical analyses~\citep{zhang2021how, carratino2022mixup, chidambaram2022towards, park2022a, pinto2022using} highlight its adaptive regularization and generalization effects, yielding stronger out-of-distribution robustness and better uncertainty calibration.
Empirical studies confirm its effectiveness in semi-supervised learning~\citep{berthelot2019mixmatch, Berthelot2020ReMixMatch, Li2020DivideMix, li2022who} and NLP~\citep{chen2020mixtext, guo2020sequence, sun2020mixup, park2022data, yang2022enhancing}.
In this paper, we extend Mixup to LLM instruction tuning, proposing a regularization method to reduce overfitting to confident examples while supporting learning for less confident ones.
% \vspace{-0.5mm}
\section{SFTMix} \label{sec:3}
% \vspace{-0.25mm}

Based on the preliminaries in \S\ref{sec:3.1}, we discuss the motivation in \S\ref{sec:3.2}, introduce SFTMix in \S\ref{sec:3.3}, and analyze its effect on the tuning process in \S\ref{sec:3.4}.

% \vspace{-0.25mm}
\subsection{Preliminaries} \label{sec:3.1}

\paragraph{The NTP Instruction-Tuning Paradigm.}
Consider an SFT dataset $\mathcal{D} = \{(\mathcal{X}_i, \mathcal{Y}_i)\}_{i=1}^{|\mathcal{D}|}$, which consists of pairs of instructions $\mathcal{X}_i$ and desired responses $\mathcal{Y}_i$.
Here, $\mathcal{X}_i = (x_1, \dots, x_{M_i})$ and $\mathcal{Y}_i = (y_1, \dots, y_{N_i})$ are sequences of tokens.
For an LLM with multiple transformer layers~\citep{vaswani2017attention} and a linear causal language modeling head $\mathbf{W}$, the conventional NTP task minimizes the following loss for predicting $\mathcal{Y}_i$ given $\mathcal{X}_i$: 
\begin{align*}
    \ell_\text{NTP}(\mathcal{D}) 
    & = - \sum_{i=1}^{|\mathcal{D}|} \sum_{n=1}^{N_i} \log p(y_n \,|\, \mathcal{X}_i, y_1, \dots, y_{n-1}) \\
    & = - \sum_{i=1}^{|\mathcal{D}|} \sum_{n=1}^{N_i} H(\mathbf{Y}_n, \sigma(\mathbf{Z}_n^\top \mathbf{W})) \text{.}
\end{align*}
This loss equals the sum of negative cross-entropy $H$ between $\mathbf{Y}_n$ and $\mathbf{Z}_n^\top \mathbf{W}$ after softmax $\sigma$, where $\mathbf{Y}_n$ is the one-hot encoding of the $n$-th token in $\mathcal{Y}_i$.
The corresponding representation $\mathbf{Z}_n$ is the last hidden state from the LLM's transformer layers: $\mathbf{Z}_n = \text{Transformers}(\mathcal{X}_i, y_1, \dots, y_{n-1})$.

\paragraph{LLM Confidence via Training Dynamics.}
Suppose we collect $C$ checkpoints of an LLM when instruction-tuning it on a dataset $\mathcal{D}$ via NTP.
We can capture the training dynamics~\citep{swayamdipta2020dataset} of this LLM by computing its confidence in generating each pair $(\mathcal{X}_i, \mathcal{Y}_i) \in \mathcal{D}$.
Specifically, let $\text{Perp}_c(\mathcal{Y}_i \,|\, \mathcal{X}_i)$ denote the LLM's perplexity for an instruction-response pair $(\mathcal{X}_i, \mathcal{Y}_i)$ at checkpoint $c \in \{1,\dots, C\}$.
Since lower perplexity implies higher generation likelihood, we define its confidence in predicting $\mathcal{Y}_i$ given $\mathcal{X}_i$ as the negative average perplexity over the $C$ checkpoints:
% \vspace{-1mm}
\begin{align*}
    \text{Conf}\,(\mathcal{Y}_i \,|\, \mathcal{X}_i) 
    = -\frac{1}{C} \sum_{c=1}^C \text{Perp}_c(\mathcal{Y}_i \,|\, \mathcal{X}_i) \text{.}
\end{align*}

\begin{figure}[t!]
    \centering
    \includegraphics[width=\columnwidth]{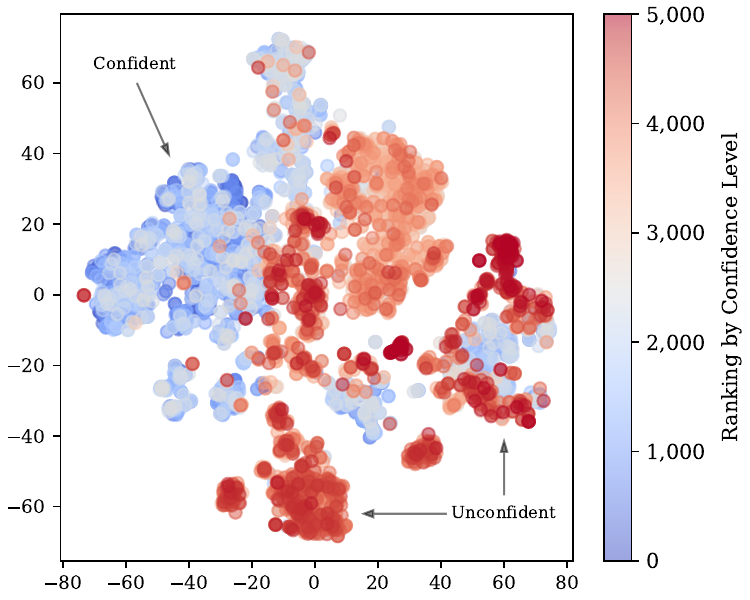}
    % \vspace{-7mm}
    \caption{Embeddings of $2{,}500$ most and $2{,}500$ least confident examples in Alpaca-52K by Llama-3.1-8B trained using NTP. The clear separation between these embeddings suggests that the LLM exhibits varying confidence levels across different semantic regions.}
    % \vspace{-2mm}
    \label{fig:conf_embs}
\end{figure}

% \vspace{-2.25mm}
\subsection{Motivation} \label{sec:3.2}

We motivate the design of SFTMix through a case study.
Specifically, we instruction-tune Llama-3.1-8B~\citep{dubey2024llama} on Alpaca-52K~\citep{alpaca} and collect the LLM's confidence for each training data point across five checkpoints.
Using the last hidden state $\mathbf{Z}$ of the final token in $(\mathcal{X}_i, \mathcal{Y}_i)$ as its semantic representation, we visualize $2{,}500$ most and $2{,}500$ least confident examples via t-SNE~\citep{van2008visualizing}.
As shown in Figure~\ref{fig:conf_embs}, embeddings of data points with contrasting confidence levels are clearly separated, indicating that \textbf{the LLM exhibits uneven confidence across the semantic representation space.}

\begin{figure*}[t!]
    \centering
    \includegraphics[width=\textwidth]{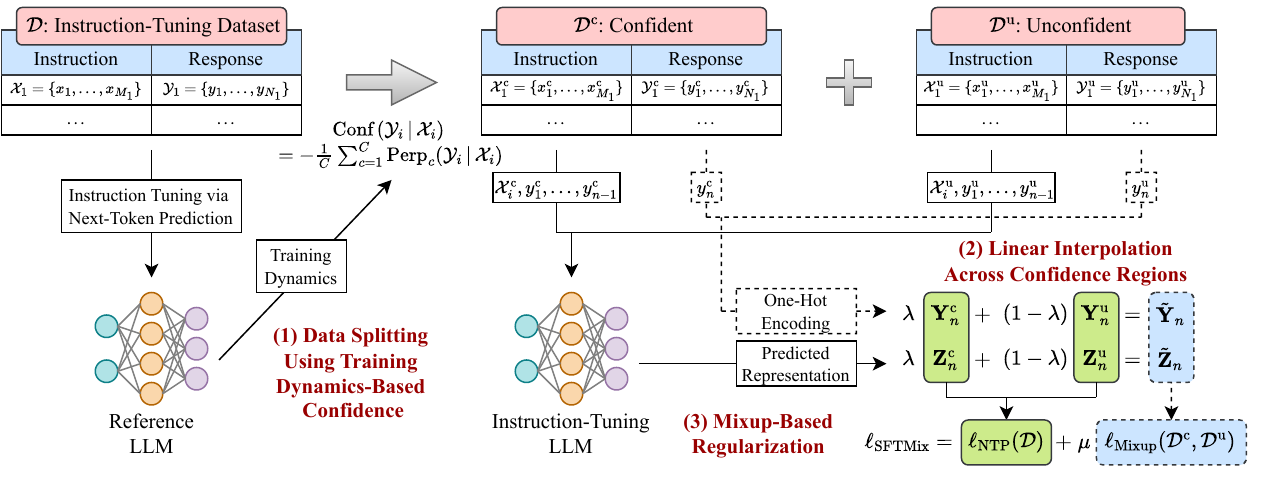}
    % \vspace{-7mm}
    \caption{The overall pipeline of the three-stage SFTMix recipe for LLM instruction tuning.}
    % \vspace{-3mm}
    \label{fig:pipeline}
\end{figure*}

Furthermore, we analyze the distributions of instruction topics in the $50$ most and $50$ least confident examples.
We find that the most confident examples primarily involve deterministic grammar tasks (e.g., ``correct any grammar error in the following sentence''), while $56\%$ of the least confident examples require creative content generation (e.g., ``find a name for an e-commerce website''), and the remaining $44\%$ consist of noisy or unanswered instructions.
This aligns with our observation in Figure~\ref{fig:conf_embs}, showing that the LLM's confidence varies across different semantic regions.

The insight from the case study motivates us to contend that \textbf{data with distinct confidence levels should play different roles during instruction tuning.}
Highly confident data points typically lie further from the classification decision boundary, posing a higher risk of overfitting~\citep{zhang2021mixup, han2024selective}.
In contrast, less confident data points are often closer to the boundary, making them harder to learn~\citep{elsayed2018large, jiang2018predicting}.

To address this, we propose SFTMix, a Mixup-based~\citep{zhang2018mixup} recipe (details in \S\ref{sec:3.3}).
Leveraging training dynamics-based confidence, we first linearly interpolate between confident and unconfident examples to bridge the confidence gap across the semantic representation space.
Then, we introduce a Mixup-based regularization to support learning on these additional, interpolated examples.
By promoting the flow of supervision signals between regions of differing confidence levels~\citep{bengio2009curriculum, chapelle2009semi} and encouraging linear behavior over a smoother decision boundary~\citep{zhang2018mixup}, our regularization mitigates overfitting in confident examples and enhances generalization in unconfident ones during LLM instruction tuning.

% \vspace{-0.5mm}
\subsection{Recipe} \label{sec:3.3}
% \vspace{-0.5mm}

We now introduce the details of our three-step SFTMix recipe (illustrated in Figure~\ref{fig:pipeline}).

\paragraph{Step 1: Determine Subspaces with Distinct Confidence Levels.}
Given an SFT dataset $\mathcal{D}$, we first instruction-tune a reference LLM via NTP and collect its confidence $\text{Conf}\,(\mathcal{Y}_i \,|\, \mathcal{X}_i)$ as in \S\ref{sec:3.1} for each pair $(\mathcal{X}_i, \mathcal{Y}_i) \in \mathcal{D}$.
We then divide $\mathcal{D}$ into two equal-sized subsets according to $\text{Conf}\,(\mathcal{Y}_i \,|\, \mathcal{X}_i)$: a confident subset $\mathcal{D}^\text{c}$ and an unconfident subset $\mathcal{D}^\text{u}$.

\paragraph{Step 2: Linearly Interpolate Confident and Unconfident Examples.}
Consider a confident instruction-response pair $(\mathcal{X}_i^\text{c}, \mathcal{Y}_i^\text{c}) \in \mathcal{D}^\text{c}$ and an unconfident pair $(\mathcal{X}_i^\text{u}, \mathcal{Y}_i^\text{u}) \in \mathcal{D}^\text{u}$.
Let $\mathbf{Y}_n^\text{c}$ and $\mathbf{Y}_n^\text{u}$ be the one-hot encoding vectors of the $n$-th token in $\mathcal{Y}^\text{c}$ and $\mathcal{Y}^\text{u}$, respectively, with $\mathbf{Z}_n^\text{c}$ and $\mathbf{Z}_n^\text{u}$ as the corresponding representations predicted by the target instruction-tuning LLM (different from the reference LLM used in Step 1).
We linearly interpolate the two pairs as follows:
% \vspace{-1mm}
\begin{align*}
    \tilde{\mathbf{Z}}_n & = \lambda \mathbf{Z}_n^\text{c} + (1-\lambda) \mathbf{Z}_n^\text{u}, \\
    \tilde{\mathbf{Y}}_n & = \lambda \mathbf{Y}_n^\text{c} + (1-\lambda) \mathbf{Y}_n^\text{u},
\end{align*}
% \vspace{-1mm}
where $\lambda \sim \text{Beta}(\alpha, \alpha)$ and $\alpha$ is a hyperparameter.

\paragraph{Step 3: Incorporate a Mixup-Based Regularization.}
Suppose that $N'_i = \min(N_i^\text{c}, N_i^\text{u})$ represents the length of the shorter response between $\mathcal{Y}_i^\text{c}$ and $\mathcal{Y}_i^\text{u}$. 
We define the Mixup-based regularization $\ell_\text{Mixup}(\mathcal{D}^\text{c}, \mathcal{D}^\text{u})$ between the confident and unconfident subsets and the overall instruction-tuning loss $\ell_\text{SFTMix}$ used in our SFTMix recipe as follows:
% \vspace{-1mm}
\begin{align*}
    & \ell_\text{Mixup}(\mathcal{D}^\text{c}, \mathcal{D}^\text{u}) = - \sum_{i=1}^{|D|/2} \sum_{n=1}^{N'_i} H(\tilde{\mathbf{Y}}_n, \sigma(\tilde{\mathbf{Z}}_n^\top \mathbf{W})) \text{,} \\
    & \ell_\text{SFTMix}(\mathcal{D}) = \ell_\text{NTP}(\mathcal{D}) + \mu \, \ell_\text{Mixup}(\mathcal{D}^\text{c}, \mathcal{D}^\text{u}) \text{.}
\end{align*}
% \vspace{-0.5mm}
Here, $\mu$ is a hyperparameter to control the regularization effect.
For ease of implementation, we ensure that each training batch contains an equal number of confident and unconfident examples, which are then randomly paired for the linear interpolation and the Mixup-based regularization.

% \vspace{-0.75mm}
\subsection{Analysis} \label{sec:3.4}
% \vspace{-0.5mm}

Here, we analyze how the Mixup-based regularization affects the direction of gradient descent in the original NTP paradigm.
For notational simplicity, we focus on the cross-entropy $H(\tilde{\mathbf{Y}}, \sigma(\tilde{\mathbf{Z}}^\top \mathbf{W}))$ between the interpolated one-hot encoding vector $\tilde{\mathbf{Y}}$ and the corresponding interpolated representation $\tilde{\mathbf{Z}}$ from the target LLM's last transformer layer. 

Let $\tilde{\mathbf{S}} = \tilde{\mathbf{Z}}^\top \mathbf{W}$, the gradient of $H$ w.r.t. $\tilde{\mathbf{S}}$ is
\begin{align*}
    \nabla_{\tilde{\mathbf{S}}} H(\tilde{\mathbf{Y}}, \sigma(\tilde{\mathbf{S}})) = \sigma(\tilde{\mathbf{S}}) - \tilde{\mathbf{Y}}.
\end{align*}
Using the chain rule, we have
\begin{align*}
    \nabla_{\mathbf{W}} H(\tilde{\mathbf{Y}}, \sigma(\tilde{\mathbf{Z}}^\top \mathbf{W})) = \tilde{\mathbf{Z}}^\top \left( \sigma(\tilde{\mathbf{Z}}^\top \mathbf{W}) - \tilde{\mathbf{Y}} \right).
\end{align*}
Since the gradient w.r.t. $\mathbf{W}$ involves the nonlinear softmax operation $\sigma$, we have
\begin{align*}
    \sigma(\tilde{\mathbf{Z}}^\top \mathbf{W})
    & = \sigma(\lambda \mathbf{Z}^{\text{c}\top} \mathbf{W} + (1-\lambda) \mathbf{Z}^{\text{u}\top} \mathbf{W}) \\
    & \neq \lambda \sigma(\mathbf{Z}^{\text{c}\top} \mathbf{W}) + (1-\lambda) \sigma(\mathbf{Z}^{\text{u}\top} \mathbf{W}).
\end{align*}
In other words, the gradient from the regularization with the interpolated example does not decompose into a weighted sum of the gradients from the NTP loss on the corresponding confident and unconfident examples. 
As a result, the Mixup-based regularization modifies the gradient descent direction in NTP by incorporating these interpolated examples.
\section{Experiments} \label{sec:4}

We assess the effectiveness of SFTMix against the NTP baseline in both instruction-following (\S\ref{sec:4.1}) and domain-specific (\S\ref{sec:4.2}) SFT tasks.

\subsection{Instruction-Following SFT} \label{sec:4.1}

\begin{table}[t!]
    \centering
    \setlength\tabcolsep{3pt}
\resizebox{\columnwidth}{!}{
    \begin{tabular}{cc|ccc|cc}
        \toprule
            \multirow{2}{*}{\textbf{LLM}} & \multirow{2}{*}{\textbf{Recipe}} & \multicolumn{3}{c|}{\textbf{MT-Bench}} & \multicolumn{2}{c}{\textbf{AlpacaEval-2}} \\
             & & \textbf{ST} & \textbf{MT} & \textbf{Overall} & \textbf{WR} & \textbf{LC WR} \\
        \midrule
        \midrule
            \multicolumn{7}{l}{\textbf{Dataset:} Alpaca-52K} \\
        \midrule
            \multirow{3}{*}{Llama} & IR-DRO & $4.8503$ & $3.6121$ & $4.2312$ & $4.1019$ & $8.7509$ \\
            & NTP & $4.9100$ & $3.8150$ & $4.3625$ & $4.0714$ & $8.6528$ \\
            & SFTMix & $\textbf{5.2125}$ & $\textbf{3.9525}$ & $\textbf{4.5825}$ & $\textbf{4.9031}$ & $\textbf{10.3195}$ \\
        \midrule
            \multirow{3}{*}{Mistral} & IR-DRO & $5.1127$ & $4.0522$ & $4.5825$ & $4.3411$ & $9.2137$ \\
            & NTP & $5.1650$ & $4.0675$ & $4.6163$ & $4.3560$ & $9.1759$ \\
            & SFTMix & $\textbf{5.2775}$ & $\textbf{4.5425}$ & $\textbf{4.9100}$ & $\textbf{4.5386}$ & $\textbf{9.4994}$ \\
        \midrule
            \multirow{2}{*}{Qwen} & NTP & $6.8177$ & $5.5683$ & $6.1930$ & $7.0764$ & $13.9508$ \\
            & SFTMix & $\textbf{7.1298}$ & $\textbf{5.9196}$ & $\textbf{6.5247}$ & $\textbf{7.8810}$ & $\textbf{15.0235}$ \\
        \midrule
            \multicolumn{7}{l}{\textbf{Dataset:} UltraChat-200K} \\
        \midrule
            \multirow{2}{*}{Llama} & NTP & $6.1875$ & $5.0125$ & $5.6000$ & $5.0665$ & $8.4505$ \\
            & SFTMix & $\textbf{6.2750}$ & $\textbf{5.3500}$ & $\textbf{5.8125}$ & $\textbf{5.1149}$ & $\textbf{9.3810}$ \\
        \midrule
            \multirow{2}{*}{Mistral} & NTP & $5.7625$ & $4.6938$ & $5.2281$ & $4.4899$ & $7.7732$ \\
            & SFTMix & $\textbf{5.9813}$ & $\textbf{4.8813}$ & $\textbf{5.4313}$ & $\textbf{4.6117}$ & $\textbf{8.7650}$ \\
        \midrule
            \multicolumn{7}{l}{\textbf{Dataset:} Tulu3-939K} \\
        \midrule
            \multirow{2}{*}{Llama} & NTP & $6.2500$ & $4.9625$ & $5.6063$ & $7.1045$ & $12.0811$ \\
            & SFTMix & $\textbf{6.6721}$ & $\textbf{5.3331}$ & $\textbf{6.0026}$ & $\textbf{8.0129}$ & $\textbf{13.1345}$ \\ 
        \midrule
            \multirow{2}{*}{Mistral} & NTP & $5.9246$ & $4.0128$ & $4.9687$ & $6.9926$ & $11.2471$ \\
            & SFTMix & $\textbf{6.3626}$ & $\textbf{4.3398}$ & $\textbf{5.3512}$ & $\textbf{7.5421}$ & $\textbf{11.8681}$ \\ 
        \bottomrule
    \end{tabular}
}
    % \vspace{-1.25mm}
    \caption{Evaluation of instruction-following capabilities of LLMs trained with NTP, SFTMix, and IR-DRO. We highlight the best-performing instruction-tuning recipe in bold. SFTMix outperforms the baselines consistently across instruction-tuning datasets and LLMs.}
    % \vspace{-1mm}
    \label{tab:ins_sft}
\end{table}

\paragraph{Experiment Setup.}
We compare SFTMix with NTP and a sample-reweighting baseline, IR-DRO~\citep{chen2024take}, using three pre-trained LLMs: Llama-3.1-8B (Llama)~\citep{dubey2024llama}, Mistral-7B-v0.1 (Mistral)~\citep{jiang2023mistral}, and Qwen-2.5-14B (Qwen)~\citep{hui2024qwen25}.
Due to computational constraints, we train Qwen only on the smaller, uncurated Alpaca-52K~\citep{alpaca}, while Llama and Mistral are trained on both Alpaca-52K and the larger filtered datasets: UltraChat-200K~\citep{tunstall2023zephyr} and Tulu3-939K~\citep{lambert2024tulu}.
For the same reason, IR-DRO is evaluated only on Llama and Mistral using Alpaca-52K.
We then evaluate the instruction-tuned LLMs on two instruction-following benchmarks: MT-Bench~\citep{zheng2023judging} and AlpacaEval-2~\citep{dubois2024lengthcontrolled}.
Following~\citet{zhao2024long}, we also conduct a human evaluation for head-to-head comparisons using the Vicuna subset in AlpacaEval-2.

\paragraph{Implementation Details.}
By default, we use a separate instance of the same type as the target instruction-tuning LLM to obtain $\text{Conf}\,(\mathcal{Y}_i \,|\, \mathcal{X}_i)$ in Step 1 of SFTMix.
We train each LLM on Alpaca-52K for three epochs and on UltraChat-200K and Tulu3-939K for one epoch, with a batch size of $32$ on eight H100 GPUs.
The tuning process employs the AdamW optimizer with a learning rate of $2\mathrm{e}{-6}$, a weight decay of $0.1$, and a cosine learning rate scheduler with a warm-up ratio of $0.1$.
Based on our hyperparameter search in \S\ref{app:a.1}, we set $\alpha = 0.5$ for sampling $\lambda$ and $\mu=0.2$ when constructing $\ell_\text{SFTMix}$.
The NTP baseline follows the same setup but excludes the Mixup-based regularization $\ell_\text{Mixup}$.
When training on UltraChat-200K and Tulu3-939K, we expand each multi-turn interaction into multiple single-turn interactions by incorporating the chat history into the instructions.
In MT-Bench and AlpacaEval-2, we employ GPT-4~\citep{achiam2023gpt} for LLM-as-a-judge and report the results averaged over five evaluation rounds.

\paragraph{Evaluation Results.}
As illustrated in Table~\ref{tab:ins_sft}, instruction-tuning with SFTMix consistently outperforms NTP and IR-DRO across all metrics in both evaluation benchmarks, regardless of the base LLM or SFT dataset. 
Notably, SFTMix yields a greater improvement in the multi-turn (MT) conversational ability (an average increase of $0.32$) compared to single-turn (ST) performance (an average increase of $0.27$) in MT-Bench.  
In AlpacaEval-2, the improvement is particularly evident in the length-controlled win rate~\citep{dubois2024lengthcontrolled} (LC WR), which better aligns with human judgment by adjusting for GPT-4's preference for longer responses. 
While instruction-tuning with the larger, higher-quality UltraChat-200K dataset results in higher scores in MT-Bench and raw win rates (WRs) in AlpacaEval-2, it also produces longer responses, leading to relatively lower LC WRs.
In contrast, SFTMix with Tulu3-939K delivers larger gains across all metrics than UltraChat-200K.

\paragraph{Further Analysis.}
Our human evaluation indicates that instruction-tuning with SFTMix wins $42.5\%$ of the head-to-head comparisons, while NTP wins only $26.5\%$ (details in \S\ref{app:a.2}).
This agrees with the conclusion from LLM-as-a-judge evaluations.
We also provide per-category scores and qualitative examples from MT-Bench in \S\ref{app:a.3} to illustrate these differences and demonstrate the effectiveness of SFTMix in multilingual instruction tuning in \S\ref{app:a.4}.
Furthermore, we compare the confidence distribution of Llama instruction-tuned with SFTMix versus NTP on Alpaca-52K. 
SFTMix reduces the standard deviation of confidence scores by $7\%$, indicating a more uniform distribution.
This suggests SFTMix helps prevent overfitting on confident samples while better supporting learning from less confident ones.

\begin{table}[t!]
    \centering
    \setlength\tabcolsep{3pt}
\resizebox{\columnwidth}{!}{
    \begin{tabular}{l|cccc|c}
        \toprule
            \multirow{2}{*}{\textbf{LLM}} & \textbf{Med} & \textbf{Med} & \textbf{PubMed} & \textbf{MedMC} & \multirow{2}{*}{\textbf{Ave}} \\
            & \textbf{QA} & \textbf{QA-5} & \textbf{QA} & \textbf{QA} & \\
        \midrule
        \midrule
            \multicolumn{6}{l}{Existing 7B Biomedical LLMs} \\
        \midrule
            MedAlpaca & 38.94 & 33.96 & 57.20 & 34.90 & 41.25 \\
            PMC-LLaMA & 27.94 & 21.24 & 54.87 & 24.57 & 32.16 \\
            BioMedGPT & 38.62 & 34.72 & 58.27 & 35.57 & 41.80 \\
            Meditron & 35.09 & 26.73 & 56.93 & 34.03 & 38.20 \\
            BioMistral & 43.86 & 37.58 & 50.13 & 44.14 & 43.93 \\
        \midrule
            \multicolumn{6}{l}{\textbf{Dataset:} MedAlpaca-263K} \\
        \midrule
            Llama & 59.68 & 53.23 & 73.40 & 52.79 & 59.78 \\
            + NTP & 59.31 & 54.52 & 75.40 & 53.65 & 60.72 \\
            or SFTMix & \textbf{60.88} & \textbf{55.38} & \textbf{77.80} & \textbf{54.15} & \textbf{62.05} \\
        \midrule
            Mistral & 49.18 & 43.94 & 72.33 & 47.98 & 53.36 \\
            + NTP & 49.10 & 44.62 & 75.40 & 48.15 & 54.32 \\
            or SFTMix & \textbf{51.77} & \textbf{45.72} & \textbf{77.40} & \textbf{49.03} & \textbf{55.98} \\
        \bottomrule
    \end{tabular}
}
    % \vspace{-2mm}
    \caption{Evaluation results on four healthcare-related benchmarks by prior biomedical LLMs and LLMs trained using either NTP or SFTMix. We bold the scores from the best-performing instruction-tuning recipe. SFTMix achieves a $1.5\%$ absolute increase in average accuracy compared to NTP for both Llama and Mistral.}
    % \vspace{-1mm}
    \label{tab:domain_sft}
\end{table}

\begin{table}[t!]
    \centering
    \setlength\tabcolsep{5pt}
\resizebox{\columnwidth}{!}{
    \begin{tabular}{c|ccc|cc}
        \toprule
            \textbf{Reference} & \multicolumn{3}{c|}{\textbf{MT-Bench}} & \multicolumn{2}{c}{\textbf{AlpacaEval-2}} \\
            \textbf{LLM} & \textbf{ST} & \textbf{MT} & \textbf{Overall} & \textbf{WR} & \textbf{LC WR} \\
        \midrule
        \midrule
            \underline{Same} & $\textbf{5.2125}$ & $3.9525$ & $\textbf{4.5825}$ & $\textbf{4.9031}$ & $\textbf{10.3195}$ \\
            Weaker & $4.8500$ & $\textbf{4.2625}$ & $4.5563$ & $4.5786$ & $10.0483$ \\            
        \bottomrule
    \end{tabular}
}
    % \vspace{-2mm}
    \caption{Evaluation of using different reference LLMs to obtain confidence in SFTMix. By default, SFTMix uses a reference LLM of the \underline{same} type as the target instruction-tuning LLM, while ``Weaker'' refers to using a less capable reference LLM. Generalizing training dynamics from a weaker reference LLM performs comparably to using the same reference LLM.}
    % \vspace{-1mm}
    \label{tab:weak}
\end{table}

% \vspace{-1.25mm}
\subsection{Domain-Specific SFT} \label{sec:4.2}
% \vspace{-0.5mm}

\paragraph{Experiment Setup.}
In healthcare-specific SFT, we train Llama and Mistral on the MedAlpaca-263K medical dataset~\citep{han2023medalpaca} using either NTP or SFTMix for two epochs, keeping other hyperparameters as in \S\ref{sec:4.1}.
We assess their performance on four healthcare-related question-answering benchmarks: MedQA~\citep{jin2021disease}, its five-option variant MedQA-5, PubMedQA~\citep{jin2019pubmedqa}, and MedMCQA~\citep{pal2022medmcqa}.
Following~\citet{labrak2024biomistral}, we adopt a three-shot setting and report the mean accuracy over three evaluation runs. 
For comparison, we also include prior biomedical LLMs of similar size: MedAlpaca-7B~\citep{han2023medalpaca}, PMC-LLaMA-7B~\citep{wu2024pmc}, BioMedGPT-LM-7B~\citep{luo2023biomedgpt}, Meditron-7B~\citep{chen2023meditron}, and BioMistral-7B~\citep{labrak2024biomistral}.

\paragraph{Evaluation Results.}
Table~\ref{tab:domain_sft} shows that SFTMix consistently surpasses NTP across all benchmarks for both LLMs. 
In particular, SFTMix leads to a $1.33\%$ absolute improvement (from $60.72\%$ to $62.05\%$) for Llama and a $1.66\%$ increase (from $54.32\%$ to $55.98\%$) for Mistral in average accuracy across the four benchmarks. 
These models also significantly outperform existing biomedical LLMs across all benchmarks by a clear margin.
% \vspace{-1mm}
\section{Analysis} \label{sec:5}
% \vspace{-0.5mm}
Building on SFTMix's effectiveness in \S\ref{sec:4}, we analyze SFTMix in depth across six directions by instruction-tuning Llama on Alpaca-52K.

% \vspace{-0.5mm}
\subsection{Generalizing the Training Dynamics from a Weaker Reference LLM Is Feasible} \label{sec:5.1}
Inspired by~\citet{burns2024weaktostrong}, we explore whether training dynamics from a weaker reference LLM transfer to a stronger one.
Specifically, we use Gemma-2B~\citep{team2024gemma} to split Alpaca-52K into confident and unconfident subsets, then use them for Mixup regularization when tuning Llama.

In Table~\ref{tab:weak}, this alternative approach yields comparable scores on MT-Bench and AlpacaEval-2 to the default SFTMix recipe, which uses the same LLM for both training dynamics and Mixup-based instruction tuning. 
This finding aligns with the weak-to-strong generalization~\citet{burns2024weaktostrong} and suggests that confidence scores precomputed from weaker LLMs can be reused for stronger models, amortizing the computational cost over time.

\begin{table}[t!]
    \centering
    \setlength\tabcolsep{3pt}
\resizebox{\columnwidth}{!}{
    \begin{tabular}{cc|ccc|cc}
        \toprule
            \textbf{NTP Data} & \textbf{Mixup} & \multicolumn{3}{c|}{\textbf{MT-Bench}} & \multicolumn{2}{c}{\textbf{AlpacaEval-2}} \\
            \textbf{Quality} & \textbf{Included?} & \textbf{ST} & \textbf{MT} & \textbf{Overall} & \textbf{WR} & \textbf{LC WR} \\
        \midrule
        \midrule
            High & No & $\textbf{6.1175}$ & $\textbf{5.2575}$ & $\textbf{5.6875}$ & $\textbf{7.2636}$ & $11.4490$ \\
            High $+$ Low & No & $5.9000$ & $5.1825$ & $5.5412$ & $6.5871$ & $\textbf{11.9590}$ \\
            High $+$ Low & Yes & $5.8025$ & $5.0975$ & $5.4500$ & $5.9382$ & $11.1768$ \\
        \bottomrule
    \end{tabular}
}
    % \vspace{-2mm}
    \caption{Evaluation of performing Mixup based on known data quality. ``High'' refers to the higher-quality examples from GPT-4, while ``Low'' refers to the lower-quality original examples in Alpaca-52K. Simply applying Mixup regularization between these subsets does not necessarily improve performance further.}
    % \vspace{-5mm}
    \label{tab:quality}
\end{table}

\begin{figure}[t!]
    \centering    
    \includegraphics[width=\columnwidth]{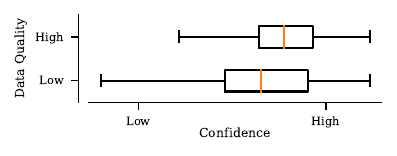}
    % \vspace{-9mm}
    \caption{Confidence distributions from instruction-tuning Llama on datasets of varying qualities. On the y-axis, ``High'' represents higher-quality examples from GPT-4, while ``Low'' denotes lower-quality original examples from Alpaca-52K. Llama's confidence distributions show substantial overlap across these datasets.}
    % \vspace{-3mm}
    \label{fig:overlap_conf}
\end{figure}

\begin{table}[t!]
    \centering
    \setlength\tabcolsep{3pt}
\resizebox{\columnwidth}{!}{
    \begin{tabular}{cc|ccc|cc}
        \toprule
            \multicolumn{2}{c|}{\textbf{Role of}} & \multicolumn{3}{c|}{\textbf{MT-Bench}} & \multicolumn{2}{c}{\textbf{AlpacaEval-2}} \\
            $\ell_\text{NTP}$ & $\ell_\text{Mixup}$ & \textbf{ST} & \textbf{MT} & \textbf{Overall} & \textbf{WR} & \textbf{LC WR} \\
        \midrule
        \midrule
            Loss & - & $4.9100$ & $3.8150$ & $4.3625$ & $4.0714$ & $8.6528$ \\
            \underline{Loss} & \underline{Reg.} & $\textbf{5.2125}$ & $3.9525$ & $\textbf{4.5825}$ & $\textbf{4.9031}$ & $\textbf{10.3195}$ \\
            Loss & Loss & $4.7050$ & $\textbf{4.1075}$ & $4.4062$ & $3.9450$ & $8.2856$ \\
            - & Loss & $5.0125$ & $4.0000$ & $4.5062$ & $3.5821$ & $7.2964$ \\
        \bottomrule
    \end{tabular}
}
    % \vspace{-2.5mm}
    \caption{Evaluation of the optimal role of $\ell_\text{Mixup}$ alongside $\ell_\text{NTP}$. By default, SFTMix incorporates $\ell_\text{Mixup}$ as a \underline{regularization} together with the NTP \underline{loss} $\ell_\text{NTP}$. This setting achieves the highest scores across most metrics.}
    % \vspace{-2.25mm}
    \label{tab:regularization}
\end{table}

\begin{table}[t!]
    \centering
    \setlength\tabcolsep{3pt}
\resizebox{\columnwidth}{!}{
    \begin{tabular}{cc|ccc|cc}
        \toprule
            \textbf{NTP} & \textbf{Mixup} & \multicolumn{3}{c|}{\textbf{MT-Bench}} & \multicolumn{2}{c}{\textbf{AlpacaEval-2}} \\
            \textbf{Dataset} & \textbf{Included?} & \textbf{ST} & \textbf{MT} & \textbf{Overall} & \textbf{WR} & \textbf{LC WR} \\
        \midrule
        \midrule
            \multirow{2}{*}{\underline{Full}} & \underline{Yes} & $\textbf{5.2125}$ & $\textbf{3.9525}$ & $\textbf{4.5825}$ & $\textbf{4.9031}$ & $\textbf{10.3195}$ \\
            & No & $4.9100$ & $3.8150$ & $4.3625$ & $4.0714$ & $8.6528$ \\
        \midrule
            \multirow{2}{*}{Conf.} & Yes & $\textbf{4.9775}$ & $\textbf{4.1075}$ & $\textbf{4.5425}$ & $\textbf{4.4496}$ & $\textbf{9.7824}$ \\
            & No & $4.7620$ & $3.8206$ & $4.2913$ & $3.9012$ & $8.0425$ \\
        \midrule
            \multirow{2}{*}{Unconf.} & Yes & $\textbf{5.1800}$ & $\textbf{3.9050}$ & $\textbf{4.5425}$ & $\textbf{4.2030}$ & $\textbf{8.9392}$ \\
            & No & $4.7164$ & $3.8392$ & $4.2778$ & $3.6552$ & $7.9889$ \\
        \bottomrule
    \end{tabular}
}
    % \vspace{-2.25mm}
    \caption{Evaluation of using the confident subset, the unconfident subset, or the full dataset for NTP. By default, SFTMix applies $\ell_\text{NTP}$ to the \underline{full} dataset alongside $\ell_\text{Mixup}$. This setting achieves the best results among the variants, demonstrating SFTMix's effectiveness in leveraging a larger set of training examples.}
    % \vspace{-2.5mm}
    \label{tab:half}
\end{table}

% \vspace{-0.5mm}
\subsection{Training Dynamics-Based Confidence Is Crucial for Performing Mixup} \label{sec:5.2}
We now explore whether we can substitute training dynamics-based confidence with known data quality. 
To test this hypothesis, we replace half of the original responses in Alpaca-52K with higher-quality GPT-4-generated versions~\citep{peng2023instruction}, forming the ``High'' subset, while referring to the remaining lower-quality original responses as ``Low''.
We then train Llama using three approaches: (1) NTP on High, (2) NTP on the combined High $+$ Low dataset, and (3) NTP on High $+$ Low with the Mixup regularization applied between them.

The use of higher-quality responses from GPT-4 indeed enhances instruction-tuning performance on both MT-Bench and AlpacaEval-2, as shown in Table~\ref{tab:quality}. 
However, simply applying Mixup between the two datasets of varying quality does not necessarily improve performance further, as indicated by the drop in the overall MT-Bench score from $5.5412$ to $5.4500$ and the LC WR in AlpacaEval-2 from $11.9590$ to $11.1768$. 
To investigate this observation, we plot the LLM's confidence distributions for both the High and Low subsets in Figure~\ref{fig:overlap_conf}. 
The substantial overlap in confidence distributions suggests that data quality does not necessarily correlate with training dynamics-based confidence. 
This highlights the importance of training dynamics in determining the model-specific role of data points, which is crucial for effectively applying SFTMix.

% \vspace{-0.75mm}
\subsection{Incorporating Mixup as a Regularization Is More Effective} \label{sec:5.3}
% \vspace{-0.25mm}

To fully explore the effect of our proposed Mixup regularization $\ell_\text{Mixup}$, we experiment two alternative treatments: (1) treating $\ell_\text{Mixup}$ as an additional loss alongside $\ell_\text{NTP}$ (i.e., $\ell = \ell_\text{NTP} + \ell_\text{Mixup}$), rather than as a regularization; and (2) minimizing only $\ell_\text{Mixup}$ without $\ell_\text{NTP}$ (i.e., $\ell = \ell_\text{Mixup}$).

Table~\ref{tab:regularization} shows that these two variants achieve higher scores on MT-Bench but perform worse on AlpacaEval-2 compared to the NTP baseline (i.e., using only the NTP loss).
Furthermore, our SFTMix recipe, which employs $\ell_\text{Mixup}$ as a regularization together with $\ell_\text{NTP}$, still outperforms both variants across both benchmarks. 
This finding highlights the importance of incorporating the traditional NTP task during SFT and supports the conclusion that Mixup is more effective when used as a regularization alongside the standard cross-entropy loss in LLM instruction tuning.

\begin{table}[t!]
    \centering
    \setlength\tabcolsep{3pt}
\resizebox{\columnwidth}{!}{
    \begin{tabular}{cc|ccc|cc}
        \toprule
            \textbf{Data} & \multirow{2}{*}{\textbf{Recipe}} & \multicolumn{3}{c|}{\textbf{MT-Bench}} & \multicolumn{2}{c}{\textbf{AlpacaEval-2}} \\
            \textbf{Selection} & & \textbf{ST} & \textbf{MT} & \textbf{Overall} & \textbf{WR} & \textbf{LC WR} \\
        \midrule
        \midrule
            \multirow{2}{*}{AlpaGasus} & NTP & $4.9787$ & $3.5275$ & $4.2531$ & $4.0752$ & $8.7182$ \\
            & SFTMix & $\textbf{5.1725}$ & $\textbf{3.9663}$ & $\textbf{4.5694}$ & $\textbf{4.9006}$ & $\textbf{10.3089}$\\
        \midrule
            \multirow{2}{*}{Long} & NTP & $4.9338$ & $3.8936$ & $4.4137$ & $4.2691$ & $8.8523$ \\
            & SFTMix & $\textbf{5.3162}$ & $\textbf{3.9262}$ & $\textbf{4.6212}$ & $\textbf{5.0230}$ & $\textbf{10.4514}$ \\
        \midrule
            \multirow{2}{*}{Uncurated} & NTP & $4.9100$ & $3.8150$ & $4.3625$ & $4.0714$ & $8.6528$ \\
            & SFTMix & $\textbf{5.2125}$ & $\textbf{3.9525}$ & $\textbf{4.5825}$ & $\textbf{4.9031}$ & $\textbf{10.3195}$ \\
        \bottomrule
    \end{tabular}
}
    % \vspace{-2.5mm}
    \caption{Evaluation of SFTMix's compatibility with data selection methods. SFTMix seamlessly integrates with them to further enhance LLM instruction tuning.}
    % \vspace{-3.5mm}
    \label{tab:selection}
\end{table}

% \vspace{-1mm}
\subsection{SFTMix Effectively Utilizes Entire Instruction-Tuning Datasets} \label{sec:5.4}
% \vspace{-0.5mm}

As part of our SFTMix recipe, we apply the NTP loss $\ell_\text{NTP}$ to the full SFT dataset. 
Here, we consider variants where $\ell_\text{NTP}$ is applied selectively to either the confident or unconfident halves of the dataset, with or without the Mixup regularization $\ell_\text{Mixup}$.

As shown in Table~\ref{tab:half}, both variants that apply $\ell_\text{NTP}$ to only half of the dataset while incorporating Mixup achieve the same overall score on MT-Bench. 
However, the variant applying $\ell_\text{NTP}$ to the confident subset performs better on AlpacaEval-2.
Notably, both variants---where $\ell_\text{NTP}$ is applied to only half of the dataset while including Mixup---outperform the NTP baseline that applies $\ell_\text{NTP}$ to the entire dataset without Mixup.
We attribute this improvement to the impact introduced by our Mixup regularization $\ell_\text{Mixup}$.
Nevertheless, our SFTMix recipe, which leverages the full dataset for NTP and applies $\ell_\text{Mixup}$, outperforms all these variants, demonstrating its ability to effectively utilize a larger set of potentially lower-quality training examples during instruction tuning. 

% \vspace{-0.5mm}
\subsection{SFTMix Integrates Well with Data Selection Methods} \label{sec:5.5}
% \vspace{-0.25mm}

Although SFTMix performs effectively with the uncurated Alpaca-52K in \S\ref{sec:4.1}, it can be seamlessly integrated with various data selection methods.
Here, we first select $1{,}000$ high-quality examples from Alpaca-52K using either AlpaGasus~\citep{chen2024alpagasus}, which grades responses with proprietary LLMs, or Long~\citep{zhao2024long}, which chooses the longest responses. 
We then apply either NTP or SFTMix to the selected examples.

As shown in Table~\ref{tab:selection}, applying SFTMix to the uncurated dataset outperforms NTP with either data selection strategy. 
Instruction tuning on the AlpaGasus-selected subset matches performance on the full dataset, while using the longest examples performs slightly better.
Nevertheless, combining SFTMix with either method yields substantial improvements over the NTP baseline.
This suggests that integrating SFTMix with existing data selection strategies~\citep{albalak2024survey, wang2024survey} could further enhance performance in LLM instruction tuning.

\begin{table}[t!]
    \centering
    \setlength\tabcolsep{3pt}
\resizebox{\columnwidth}{!}{
    \begin{tabular}{cc|ccc|cc}
        \toprule
            \textbf{Using} & \multirow{2}{*}{\textbf{Recipe}} & \multicolumn{3}{c|}{\textbf{MT-Bench}} & \multicolumn{2}{c}{\textbf{AlpacaEval-2}} \\
            \textbf{LoRA?} & & \textbf{ST} & \textbf{MT} & \textbf{Overall} & \textbf{WR} & \textbf{LC WR} \\
        \midrule
        \midrule
            \multirow{2}{*}{Yes} & NTP & $4.9350$ & $3.7600$ & $4.3475$ & $3.8841$ & $8.5104$ \\
            & SFTMix & $\textbf{5.3350}$ & $\textbf{3.8088}$ & $\textbf{4.5719}$ & $\textbf{4.8785}$ & $\textbf{9.8030}$ \\
        \midrule
            \multirow{2}{*}{No} & NTP & $4.9100$ & $3.8150$ & $4.3625$ & $4.0714$ & $8.6528$ \\
            & SFTMix & $\textbf{5.2125}$ & $\textbf{3.9525}$ & $\textbf{4.5825}$ & $\textbf{4.9031}$ & $\textbf{10.3195}$ \\
        \bottomrule
    \end{tabular}
}
    % \vspace{-2.5mm}
    \caption{Evaluation of SFTMix's adaptability to LoRA. SFTMix outperforms NTP when using LoRA, adapting well to compute-constrained scenarios.}
    % \vspace{-4.25mm}
    \label{tab:lora}
\end{table}

% \vspace{-0.75mm}
\subsection{SFTMix Is Compatible with Parameter-Efficient Fine-Tuning} \label{sec:5.6}
% \vspace{-0.25mm}

To enable parameter-efficient fine-tuning, we test SFTMix's compatibility with low-rank adaptation (LoRA)~\citep{hu2022lora}.
Specifically, we compare SFTMix and NTP using both LoRA and full-parameter fine-tuning, with the results in Table~\ref{tab:lora}.

Overall, LoRA performs comparably to full-parameter SFT in MT-Bench but slightly underperforms in AlpacaEval-2.
Even with LoRA-based instruction tuning, SFTMix effectively improves performance over the NTP baseline, demonstrating its adaptability to compute-constrained scenarios.
% \vspace{-1mm}
\section{Conclusion} \label{sec:6}
% \vspace{-1mm}

In this paper, we propose SFTMix, a novel recipe for elevating LLM instruction tuning.
We observe that LLMs exhibit uneven confidence distributions across the semantic space, and argue that data with different confidence levels should play distinct roles in tuning.
To this end, we partition an SFT dataset into confident and unconfident subsets, interpolate them to bridge the confidence gap, and introduce a Mixup-based regularization to facilitate learning.
Extensive experiments show that SFTMix outperforms the conventional NTP paradigm across diverse LLM families and SFT datasets.
Our analyses further highlight its versatility and scalability.
Applying dynamic scheduling to Mixup regularization and extending it to LLM pre-training are promising directions for future work.

% \newpage
\section*{Limitation}
Due to computational constraints, we do not apply SFTMix to LLM pre-training or instruction-tune models larger than 14B. 
We also acknowledge that SFTMix requires an initial fine-tuning round to estimate training dynamics-based confidence. 
While this adds computational cost, it is comparable to prior data selection methods such as LESS~\citep{xia2024less} and Rho-1~\citep{lin2024not}, which also depend on reference models. 
To mitigate this cost, we explore using confidence scores from weaker LLMs in \S\ref{sec:5.1}, though further experiments are needed to robustly establish their generalization. 
We also examine SFTMix's compatibility with lightweight data selection methods in \S\ref{sec:5.5}, but additional evaluation is required to validate its effectiveness across varied strategies.

% \newpage
\bibliography{main}

@inproceedings{vaswani2017attention,
  title={Attention is all you need},
  author={Vaswani, Ashish and Shazeer, Noam and Parmar, Niki and Uszkoreit, Jakob and Jones, Llion and Gomez, Aidan N and Kaiser, {\L}ukasz and Polosukhin, Illia},
  booktitle={NeurIPS},
  year={2017}
}

@inproceedings{swayamdipta2020dataset,
  title={Dataset Cartography: Mapping and Diagnosing Datasets with Training Dynamics},
  author={Swayamdipta, Swabha and Schwartz, Roy and Lourie, Nicholas and Wang, Yizhong and Hajishirzi, Hannaneh and Smith, Noah A and Choi, Yejin},
  booktitle={EMNLP},
  year={2020}
}

@inproceedings{ouyang2022training,
  title={Training language models to follow instructions with human feedback},
  author={Ouyang, Long and Wu, Jeff and Jiang, Xu and Almeida, Diogo and Wainwright, Carroll L and Mishkin, Pamela and Zhang, Chong and Agarwal, Sandhini and Slama, Katarina and Ray, Alex and others},
  booktitle={NeurIPS},
  year={2022}
}

@article{zhang2023instruction,
  title={Instruction tuning for large language models: A survey},
  author={Zhang, Shengyu and Dong, Linfeng and Li, Xiaoya and Zhang, Sen and Sun, Xiaofei and Wang, Shuhe and Li, Jiwei and Hu, Runyi and Zhang, Tianwei and Wu, Fei and others},
  journal={arXiv preprint},
  year={2023}
}

@inproceedings{shi2024instruction,
  title={Instruction Tuning With Loss Over Instructions},
  author={Zhengyan Shi and Adam X. Yang and Bin Wu and Laurence Aitchison and Emine Yilmaz and Aldo Lipani},
  booktitle={NeurIPS},
  year={2024}
}

@inproceedings{jainneftune,
  title={NEFTune: Noisy Embeddings Improve Instruction Finetuning},
  author={Jain, Neel and Chiang, Ping-yeh and Wen, Yuxin and Kirchenbauer, John and Chu, Hong-Min and Somepalli, Gowthami and Bartoldson, Brian R and Kailkhura, Bhavya and Schwarzschild, Avi and Saha, Aniruddha and others},
  booktitle={ICLR},
  year={2024}
}

@inproceedings{xie2023data,
  title={Data selection for language models via importance resampling},
  author={Xie, Sang Michael and Santurkar, Shibani and Ma, Tengyu and Liang, Percy},
  booktitle={NeurIPS},
  year={2023}
}

@inproceedings{xia2024less,
   title={{LESS}: Selecting Influential Data for Targeted Instruction Tuning},
   author={Xia, Mengzhou and Malladi, Sadhika and Gururangan, Suchin and Arora, Sanjeev and Chen, Danqi},
   booktitle={ICML},
   year={2024}
}

@inproceedings{zhou2023lima,
title={{LIMA}: Less Is More for Alignment},
author={Chunting Zhou and Pengfei Liu and Puxin Xu and Srini Iyer and Jiao Sun and Yuning Mao and Xuezhe Ma and Avia Efrat and Ping Yu and LILI YU and Susan Zhang and Gargi Ghosh and Mike Lewis and Luke Zettlemoyer and Omer Levy},
booktitle={NeurIPS},
year={2023}
}

@inproceedings{chen2024alpagasus,
title={AlpaGasus: Training a Better Alpaca with Fewer Data},
author={Lichang Chen and Shiyang Li and Jun Yan and Hai Wang and Kalpa Gunaratna and Vikas Yadav and Zheng Tang and Vijay Srinivasan and Tianyi Zhou and Heng Huang and Hongxia Jin},
booktitle={ICLR},
year={2024}
}

@inproceedings{zhao2024long,
title={Long Is More for Alignment: A Simple but Tough-to-Beat Baseline for Instruction Fine-Tuning},
author={Hao Zhao and Maksym Andriushchenko and Francesco Croce and Nicolas Flammarion},
booktitle={ICML},
year={2024}
}

@inproceedings{schoch2023data,
  title={Data Selection for Fine-tuning Large Language Models Using Transferred Shapley Values},
  author={Schoch, Stephanie and Mishra, Ritwick and Ji, Yangfeng},
  booktitle={ACL Workshop},
  year={2023}
}

@inproceedings{ding2023enhancing,
  title={Enhancing Chat Language Models by Scaling High-quality Instructional Conversations},
  author={Ding, Ning and Chen, Yulin and Xu, Bokai and Qin, Yujia and Hu, Shengding and Liu, Zhiyuan and Sun, Maosong and Zhou, Bowen},
  booktitle={EMNLP},
  year={2023}
}

@inproceedings{xu2024wizardlm,
  title={WizardLM: Empowering large pre-trained language models to follow complex instructions},
  author={Xu, Can and Sun, Qingfeng and Zheng, Kai and Geng, Xiubo and Zhao, Pu and Feng, Jiazhan and Tao, Chongyang and Lin, Qingwei and Jiang, Daxin},
  booktitle={ICLR},
  year={2024}
}

@inproceedings{wang2023self,
  title={Self-Instruct: Aligning Language Models with Self-Generated Instructions},
  author={Wang, Yizhong and Kordi, Yeganeh and Mishra, Swaroop and Liu, Alisa and Smith, Noah A and Khashabi, Daniel and Hajishirzi, Hannaneh},
  booktitle={ACL},
  year={2023}
}

@misc{alpaca,
  author = {Rohan Taori and Ishaan Gulrajani and Tianyi Zhang and Yann Dubois and Xuechen Li and Carlos Guestrin and Percy Liang and Tatsunori B. Hashimoto },
  title = {Stanford Alpaca: An Instruction-following LLaMA model},
  year = {2023},
  url = {https://github.com/tatsu-lab/stanford_alpaca}
}

@misc{chen2023multilingualsift,
  title={MultilingualSIFT: Multilingual supervised instruction fine-tuning},
  author={Chen, Zhihong and Yan, Shuo and Liang, Juhao and Jiang, Feng and Wu, Xiangbo and Yu, Fei and Chen, Guiming Hardy and Chen, Junying and Zhang, Hongbo and Jianquan, Li and others},
  year={2023},
  publisher={GitHub}
}

@misc{vicuna2023,
    title = {Vicuna: An Open-Source Chatbot Impressing GPT-4 with 90\%* ChatGPT Quality},
    url = {https://lmsys.org/blog/2023-03-30-vicuna/},
    author = {Chiang, Wei-Lin and Li, Zhuohan and Lin, Zi and Sheng, Ying and Wu, Zhanghao and Zhang, Hao and Zheng, Lianmin and Zhuang, Siyuan and Zhuang, Yonghao and Gonzalez, Joseph E. and Stoica, Ion and Xing, Eric P.},
    year = {2023}
}

@inproceedings{lin2024not,
title={Not All Tokens Are What You Need for Pretraining},
author={Zhenghao Lin and Zhibin Gou and Yeyun Gong and Xiao Liu and yelong shen and Ruochen Xu and Chen Lin and Yujiu Yang and Jian Jiao and Nan Duan and Weizhu Chen},
booktitle={NeurIPS},
year={2024}
}

@inproceedings{kung2023active,
  title={Active Instruction Tuning: Improving Cross-Task Generalization by Training on Prompt Sensitive Tasks},
  author={Kung, Po-Nien and Yin, Fan and Wu, Di and Chang, Kai-Wei and Peng, Nanyun},
  booktitle={EMNLP},
  year={2023}
}

@inproceedings{rao2024commonit,
  title={CommonIT: Commonality-Aware Instruction Tuning for Large Language Models via Data Partitions},
  author={Rao, Jun and Liu, Xuebo and Lian, Lian and Cheng, Shengjun and Liao, Yunjie and Zhang, Min},
  booktitle={EMNLP},
  year={2024}
}

@article{albalak2024survey,
  title={A survey on data selection for language models},
  author={Albalak, Alon and Elazar, Yanai and Xie, Sang Michael and Longpre, Shayne and Lambert, Nathan and Wang, Xinyi and Muennighoff, Niklas and Hou, Bairu and Pan, Liangming and Jeong, Haewon and others},
  journal={arXiv preprint},
  year={2024}
}

@article{wang2024survey,
  title={A Survey on Data Selection for LLM Instruction Tuning},
  author={Wang, Jiahao and Zhang, Bolin and Du, Qianlong and Zhang, Jiajun and Chu, Dianhui},
  journal={arXiv preprint},
  year={2024}
}

@article{liu2019roberta,
  title={RoBERTa: A Robustly Optimized BERT Pretraining Approach}, 
  author={Yinhan Liu and Myle Ott and Naman Goyal and Jingfei Du and Mandar Joshi and Danqi Chen and Omer Levy and Mike Lewis and Luke Zettlemoyer and Veselin Stoyanov},
  journal={arXiv preprint},
  year={2019}
}

@inproceedings{zhang2021cartography, 
  title={Cartography Active Learning},
  author={Zhang, Mike and Plank, Barbara},
  booktitle={EMNLP Findings},
  year={2021}
}

@inproceedings{christopoulou2022training,
  title={Training Dynamics for Curriculum Learning: A Study on Monolingual and Cross-lingual NLU},
  author={Christopoulou, Fenia and Lampouras, Gerasimos and Iacobacci, Ignacio},
  booktitle={EMNLP},
  year={2022}
}

@inproceedings{poesina-etal-2024-novel,
    title = "A Novel Cartography-Based Curriculum Learning Method Applied on {R}o{NLI}: The First {R}omanian Natural Language Inference Corpus",
    author = "Poesina, Eduard  and
      Caragea, Cornelia  and
      Ionescu, Radu",
    booktitle = "ACL",
    year = "2024"
}

@inproceedings{zhang2022allsh,
  title={ALLSH: Active Learning Guided by Local Sensitivity and Hardness},
  author={Zhang, Shujian and Gong, Chengyue and Liu, Xingchao and He, Pengcheng and Chen, Weizhu and Zhou, Mingyuan},
  booktitle={NAACL Findings},
  year={2022}
}

@inproceedings{chimoto-etal-2024-critical,
    title = "Critical Learning Periods: Leveraging Early Training Dynamics for Efficient Data Pruning",
    author = "Chimoto, Everlyn  and
      Gala, Jay  and
      Ahia, Orevaoghene  and
      Kreutzer, Julia  and
      Bassett, Bruce  and
      Hooker, Sara",
    booktitle = "ACL Findings",
    year = "2024"
}

@inproceedings{he2024large,
  title={Large-scale dataset pruning with dynamic uncertainty},
  author={He, Muyang and Yang, Shuo and Huang, Tiejun and Zhao, Bo},
  booktitle={CVPR},
  year={2024}
}

@inproceedings{seedat2024curated,
title={Curated {LLM}: Synergy of {LLM}s and Data Curation for tabular augmentation in low-data regimes},
author={Nabeel Seedat and Nicolas Huynh and Boris van Breugel and Mihaela van der Schaar},
booktitle={ICML},
year={2024}
}

@inproceedings{zhang2018mixup,
  title={mixup: Beyond Empirical Risk Minimization},
  author={Zhang, Hongyi and Cisse, Moustapha and Dauphin, Yann N and Lopez-Paz, David},
  booktitle={ICLR},
  year={2018}
}

@inproceedings{verma2019manifold,
  title={Manifold mixup: Better representations by interpolating hidden states},
  author={Verma, Vikas and Lamb, Alex and Beckham, Christopher and Najafi, Amir and Mitliagkas, Ioannis and Lopez-Paz, David and Bengio, Yoshua},
  booktitle={ICML},
  year={2019}
}

@inproceedings{hendrycks2020augmix,
  title={Augmix: A simple method to improve robustness and uncertainty under data shift},
  author={Hendrycks, Dan and Mu, Norman and Cubuk, Ekin Dogus and Zoph, Barret and Gilmer, Justin and Lakshminarayanan, Balaji},
  booktitle={ICLR},
  year={2020}
}

@inproceedings{uddin2021saliencymix,
title={SaliencyMix: A Saliency Guided Data Augmentation Strategy for Better Regularization},
author={A F M Shahab Uddin and Mst. Sirazam Monira and Wheemyung Shin and TaeChoong Chung and Sung-Ho Bae},
booktitle={ICLR},
year={2021}
}

@inproceedings{choi2022tokenmixup,
title={TokenMixup: Efficient Attention-guided Token-level Data Augmentation for Transformers},
author={Hyeong Kyu Choi and Joonmyung Choi and Hyunwoo J. Kim},
booktitle={NeurIPS},
year={2022}
}

@inproceedings{zhang2021how,
title={How Does Mixup Help With Robustness and Generalization?},
author={Linjun Zhang and Zhun Deng and Kenji Kawaguchi and Amirata Ghorbani and James Zou},
booktitle={ICLR},
year={2021}
}

@inproceedings{chidambaram2022towards,
title={Towards Understanding the Data Dependency of Mixup-style Training},
author={Muthu Chidambaram and Xiang Wang and Yuzheng Hu and Chenwei Wu and Rong Ge},
booktitle={ICLR},
year={2022}
}

@article{carratino2022mixup,
  title={On mixup regularization},
  author={Carratino, Luigi and Ciss{\'e}, Moustapha and Jenatton, Rodolphe and Vert, Jean-Philippe},
  journal={JMLR},
  year={2022}
}

@inproceedings{park2022a,
title={A Unified Analysis of Mixed Sample Data Augmentation: A Loss Function Perspective},
author={Chanwoo Park and Sangdoo Yun and Sanghyuk Chun},
booktitle={NeurIPS},
year={2022}
}

@article{pinto2022using,
  title={Using mixup as a regularizer can surprisingly improve accuracy \& out-of-distribution robustness},
  author={Pinto, Francesco and Yang, Harry and Lim, Ser Nam and Torr, Philip and Dokania, Puneet},
  journal={NeurIPS},
  year={2022}
}

@article{berthelot2019mixmatch,
  title={Mixmatch: A holistic approach to semi-supervised learning},
  author={Berthelot, David and Carlini, Nicholas and Goodfellow, Ian and Papernot, Nicolas and Oliver, Avital and Raffel, Colin A},
  journal={NeurIPS},
  year={2019}
}

@inproceedings{Berthelot2020ReMixMatch,
title={ReMixMatch: Semi-Supervised Learning with Distribution Matching and Augmentation Anchoring},
author={David Berthelot and Nicholas Carlini and Ekin D. Cubuk and Alex Kurakin and Kihyuk Sohn and Han Zhang and Colin Raffel},
booktitle={ICLR},
year={2020},
}

@inproceedings{Li2020DivideMix,
title={DivideMix: Learning with Noisy Labels as Semi-supervised Learning},
author={Junnan Li and Richard Socher and Steven C.H. Hoi},
booktitle={ICLR},
year={2020}
}

@inproceedings{li2022who,
title={Who Is Your Right Mixup Partner in Positive and Unlabeled Learning},
author={Changchun Li and Ximing Li and Lei Feng and Jihong Ouyang},
booktitle={ICLR},
year={2022}
}

@inproceedings{chen2020mixtext,
  title={MixText: Linguistically-Informed Interpolation of Hidden Space for Semi-Supervised Text Classification},
  author={Chen, Jiaao and Yang, Zichao and Yang, Diyi},
  booktitle={ACL},
  year={2020}
}

@inproceedings{guo2020sequence,
  title={Sequence-Level Mixed Sample Data Augmentation},
  author={Guo, Demi and Kim, Yoon and Rush, Alexander M},
  booktitle={EMNLP},
  year={2020}
}

@inproceedings{sun2020mixup,
  title={Mixup-Transformer: Dynamic Data Augmentation for NLP Tasks},
  author={Sun, Lichao and Xia, Congying and Yin, Wenpeng and Liang, Tingting and Philip, S Yu and He, Lifang},
  booktitle={COLING},
  year={2020}
}

@inproceedings{yang2022enhancing,
title={Enhancing Cross-lingual Transfer by Manifold Mixup},
author={Huiyun Yang and Huadong Chen and Hao Zhou and Lei Li},
booktitle={ICLR},
year={2022}
}

@inproceedings{park2022data,
  title={A Data Cartography based MixUp for Pre-trained Language Models},
  author={Park, Seo Yeon and Caragea, Cornelia},
  booktitle={NAACL},
  year={2022}
}

@article{dubey2024llama,
  title={The llama 3 herd of models},
  author={Dubey, Abhimanyu and Jauhri, Abhinav and Pandey, Abhinav and Kadian, Abhishek and Al-Dahle, Ahmad and Letman, Aiesha and Mathur, Akhil and Schelten, Alan and Yang, Amy and Fan, Angela and others},
  journal={arXiv preprint},
  year={2024}
}

@article{van2008visualizing,
  title={Visualizing data using t-SNE.},
  author={Van der Maaten, Laurens and Hinton, Geoffrey},
  journal={JMLR},
  year={2008}
}

@inproceedings{bengio2009curriculum,
  title={Curriculum learning},
  author={Bengio, Yoshua and Louradour, J{\'e}r{\^o}me and Collobert, Ronan and Weston, Jason},
  booktitle={ICML},
  year={2009}
}

@article{chapelle2009semi,
  title={Semi-supervised learning (chapelle, o. et al., eds.; 2006)[book reviews]},
  author={Chapelle, Olivier and Scholkopf, Bernhard and Zien, Alexander},
  journal={IEEE Transactions on Neural Networks},
  year={2009}
}

@inproceedings{sohn2020fixmatch,
  title={FixMatch: simplifying semi-supervised learning with consistency and confidence},
  author={Sohn, Kihyuk and Berthelot, David and Li, Chun-Liang and Zhang, Zizhao and Carlini, Nicholas and Cubuk, Ekin D and Kurakin, Alex and Zhang, Han and Raffel, Colin},
  booktitle={NeurIPS},
  year={2020}
}

@article{jiang2023mistral,
  title={Mistral 7B},
  author={Jiang, Albert Q and Sablayrolles, Alexandre and Mensch, Arthur and Bamford, Chris and Chaplot, Devendra Singh and Casas, Diego de las and Bressand, Florian and Lengyel, Gianna and Lample, Guillaume and Saulnier, Lucile and others},
  journal={arXiv preprint},
  year={2023}
}

@article{tunstall2023zephyr,
  title={Zephyr: Direct distillation of lm alignment},
  author={Tunstall, Lewis and Beeching, Edward and Lambert, Nathan and Rajani, Nazneen and Rasul, Kashif and Belkada, Younes and Huang, Shengyi and von Werra, Leandro and Fourrier, Cl{\'e}mentine and Habib, Nathan and others},
  journal={arXiv preprint},
  year={2023}
}

@inproceedings{zheng2023judging,
  title={Judging LLM-as-a-judge with MT-bench and Chatbot Arena},
  author={Zheng, Lianmin and Chiang, Wei-Lin and Sheng, Ying and Zhuang, Siyuan and Wu, Zhanghao and Zhuang, Yonghao and Lin, Zi and Li, Zhuohan and Li, Dacheng and Xing, Eric P and others},
  booktitle={NeurIPS},
  year={2023}
}

@inproceedings{dubois2024lengthcontrolled,
title={Length-Controlled AlpacaEval: A Simple Debiasing of Automatic Evaluators},
author={Yann Dubois and Percy Liang and Tatsunori Hashimoto},
booktitle={COLM},
year={2024}
}

@article{achiam2023gpt,
  title={Gpt-4 technical report},
  author={Achiam, Josh and Adler, Steven and Agarwal, Sandhini and Ahmad, Lama and Akkaya, Ilge and Aleman, Florencia Leoni and Almeida, Diogo and Altenschmidt, Janko and Altman, Sam and Anadkat, Shyamal and others},
  journal={arXiv preprint},
  year={2023}
}

@article{hurst2024gpt,
  title={Gpt-4o system card},
  author={Hurst, Aaron and Lerer, Adam and Goucher, Adam P and Perelman, Adam and Ramesh, Aditya and Clark, Aidan and Ostrow, AJ and Welihinda, Akila and Hayes, Alan and Radford, Alec and others},
  journal={arXiv preprint},
  year={2024}
}

@article{han2023medalpaca,
  title={MedAlpaca--An Open-Source Collection of Medical Conversational AI Models and Training Data},
  author={Han, Tianyu and Adams, Lisa C and Papaioannou, Jens-Michalis and Grundmann, Paul and Oberhauser, Tom and L{\"o}ser, Alexander and Truhn, Daniel and Bressem, Keno K},
  journal={arXiv preprint},
  year={2023}
}

@article{jin2021disease,
  title={What disease does this patient have? a large-scale open domain question answering dataset from medical exams},
  author={Jin, Di and Pan, Eileen and Oufattole, Nassim and Weng, Wei-Hung and Fang, Hanyi and Szolovits, Peter},
  journal={Applied Sciences},
  year={2021}
}

@inproceedings{pal2022medmcqa,
  title={Medmcqa: A large-scale multi-subject multi-choice dataset for medical domain question answering},
  author={Pal, Ankit and Umapathi, Logesh Kumar and Sankarasubbu, Malaikannan},
  booktitle={CHIL},
  year={2022}
}

@inproceedings{jin2019pubmedqa,
  title={PubMedQA: A Dataset for Biomedical Research Question Answering},
  author={Jin, Qiao and Dhingra, Bhuwan and Liu, Zhengping and Cohen, William and Lu, Xinghua},
  booktitle={EMNLP},
  year={2019}
}

@inproceedings{labrak2024biomistral,
    title = "{B}io{M}istral: A Collection of Open-Source Pretrained Large Language Models for Medical Domains",
    author={Labrak, Yanis and Bazoge, Adrien and Morin, Emmanuel and Gourraud, Pierre-Antoine and Rouvier, Mickael and Dufour, Richard},
    booktitle = "ACL Findings",
    year = "2024"
}

@article{wu2024pmc,
  title={PMC-LLaMA: toward building open-source language models for medicine},
  author={Wu, Chaoyi and Lin, Weixiong and Zhang, Xiaoman and Zhang, Ya and Xie, Weidi and Wang, Yanfeng},
  journal={JAMIA},
  year={2024}
}

@article{chen2023meditron,
  title={Meditron-70b: Scaling medical pretraining for large language models},
  author={Chen, Zeming and Cano, Alejandro Hern{\'a}ndez and Romanou, Angelika and Bonnet, Antoine and Matoba, Kyle and Salvi, Francesco and Pagliardini, Matteo and Fan, Simin and K{\"o}pf, Andreas and Mohtashami, Amirkeivan and others},
  journal={arXiv preprint},
  year={2023}
}

@article{luo2023biomedgpt,
  title={Biomedgpt: Open multimodal generative pre-trained transformer for biomedicine},
  author={Luo, Yizhen and Zhang, Jiahuan and Fan, Siqi and Yang, Kai and Wu, Yushuai and Qiao, Mu and Nie, Zaiqing},
  journal={arXiv preprint},
  year={2023}
}

@inproceedings{burns2024weaktostrong,
title={Weak-to-Strong Generalization: Eliciting Strong Capabilities With Weak Supervision},
author={Collin Burns and Pavel Izmailov and Jan Hendrik Kirchner and Bowen Baker and Leo Gao and Leopold Aschenbrenner and Yining Chen and Adrien Ecoffet and Manas Joglekar and Jan Leike and Ilya Sutskever and Jeffrey Wu},
booktitle={ICML},
year={2024}
}

@article{team2024gemma,
  title={Gemma 2: Improving open language models at a practical size},
  author={Team, Gemma and Riviere, Morgane and Pathak, Shreya and Sessa, Pier Giuseppe and Hardin, Cassidy and Bhupatiraju, Surya and Hussenot, L{\'e}onard and Mesnard, Thomas and Shahriari, Bobak and Ram{\'e}, Alexandre and others},
  journal={arXiv preprint},
  year={2024}
}

@article{peng2023instruction,
  title={Instruction tuning with gpt-4},
  author={Peng, Baolin and Li, Chunyuan and He, Pengcheng and Galley, Michel and Gao, Jianfeng},
  journal={arXiv preprint},
  year={2023}
}

@inproceedings{hu2022lora,
title={Lo{RA}: Low-Rank Adaptation of Large Language Models},
author={Edward J Hu and yelong shen and Phillip Wallis and Zeyuan Allen-Zhu and Yuanzhi Li and Shean Wang and Lu Wang and Weizhu Chen},
booktitle={ICLR},
year={2022}
}

@inproceedings{jiang2018predicting,
  title={Predicting the Generalization Gap in Deep Networks with Margin Distributions},
  author={Jiang, Yiding and Krishnan, Dilip and Mobahi, Hossein and Bengio, Samy},
  booktitle={ICLR},
  year={2018}
}

@inproceedings{elsayed2018large,
  title={Large margin deep networks for classification},
  author={Elsayed, Gamaleldin F and Krishnan, Dilip and Mobahi, Hossein and Regan, Kevin and Bengio, Samy},
  booktitle={NeurIPS},
  year={2018}
}

@article{han2024selective,
  title={Selective learning: Towards robust calibration with dynamic regularization},
  author={Han, Zongbo and Yang, Yifeng and Zhang, Changqing and Zhang, Linjun and Zhou, Joey Tianyi and Hu, Qinghua},
  journal={arXiv preprint},
  year={2024}
}

@article{zhang2021mixup,
  title={Mixup training leads to reduced overfitting and improved calibration for the transformer architecture},
  author={Zhang, Wancong and Vaidya, Ieshan},
  journal={arXiv preprint},
  year={2021}
}

@article{zhao2023survey,
  title={A survey of large language models},
  author={Zhao, Wayne Xin and Zhou, Kun and Li, Junyi and Tang, Tianyi and Wang, Xiaolei and Hou, Yupeng and Min, Yingqian and Zhang, Beichen and Zhang, Junjie and Dong, Zican and others},
  journal={arXiv preprint},
  year={2023}
}

@article{minaee2024large,
  title={Large language models: A survey},
  author={Minaee, Shervin and Mikolov, Tomas and Nikzad, Narjes and Chenaghlu, Meysam and Socher, Richard and Amatriain, Xavier and Gao, Jianfeng},
  journal={arXiv preprint},
  year={2024}
}

@article{hui2024qwen25,
  title={Qwen2. 5-coder technical report},
  author={Hui, Binyuan and Yang, Jian and Cui, Zeyu and Yang, Jiaxi and Liu, Dayiheng and Zhang, Lei and Liu, Tianyu and Zhang, Jiajun and Yu, Bowen and Lu, Keming and others},
  journal={arXiv preprint},
  year={2024}
}

@article{chen2024take,
  title={Take the bull by the horns: Hard sample-reweighted continual training improves llm generalization},
  author={Chen, Xuxi and Wang, Zhendong and Sow, Daouda and Yang, Junjie and Chen, Tianlong and Liang, Yingbin and Zhou, Mingyuan and Wang, Zhangyang},
  journal={arXiv preprint},
  year={2024}
}

@inproceedings{yang2025measuring,
  title={Measuring data diversity for instruction tuning: A systematic analysis and a reliable metric},
  author={Yang, Yuming and Nan, Yang and Ye, Junjie and Dou, Shihan and Wang, Xiao and Li, Shuo and Lv, Huijie and Wu, Mingqi and Gui, Tao and Zhang, Qi and others},
  booktitle={ACL},
  year={2025}
}

@inproceedings{zhao2025beyond,
  title={Beyond similarity: A gradient-based graph method for instruction tuning data selection},
  author={Zhao, Yang and Du, Li and Ding, Xiao and Ouyang, Yangou and Wang, Hepeng and Xiong, Kai and Gao, Jinglong and Sun, Zhouhao and Xu, Dongliang and Qing, Yang and others},
  booktitle={ACL},
  year={2025}
}

@inproceedings{zhang2025best,
  title={The best instruction-tuning data are those that fit},
  author={Zhang, Dylan and Dai, Qirun and Peng, Hao},
  booktitle={NeurIPS},
  year={2025}
}

@inproceedings{fu2025t,
  title={T-SHIRT: Token-Selective Hierarchical Data Selection for Instruction Tuning},
  author={Fu, Yanjun and Hamman, Faisal and Dutta, Sanghamitra},
  booktitle={NeurIPS},
  year={2025}
}

@inproceedings{wang2025nice,
title={{NICE} Data Selection for Instruction Tuning in {LLM}s with Non-differentiable Evaluation Metric},
author={Jingtan Wang and Xiaoqiang Lin and Rui Qiao and Pang Wei Koh and Chuan-Sheng Foo and Bryan Kian Hsiang Low},
booktitle={ICML},
year={2025}
}

@inproceedings{qi2025evolm,
  title={EvoLM: In Search of Lost Language Model Training Dynamics},
  author={Qi, Zhenting and Nie, Fan and Alahi, Alexandre and Zou, James and Lakkaraju, Himabindu and Du, Yilun and Xing, Eric and Kakade, Sham and Zhang, Hanlin},
  booktitle={NeurIPS},
  year={2025}
}

@inproceedings{zhang2025training,
title={Training Dynamics of In-Context Learning in Linear Attention},
author={Yedi Zhang and Aaditya K Singh and Peter E. Latham and Andrew M Saxe},
booktitle={ICML},
year={2025}
}

@inproceedings{mircea2025training,
  title={Training Dynamics Underlying Language Model Scaling Laws: Loss Deceleration and Zero-Sum Learning},
  author={Mircea, Andrei and Chakraborty, Supriyo and Chitsazan, Nima and Naphade, Milind and Sahu, Sambit and Rish, Irina and Lobacheva, Ekaterina},
  booktitle={ACL},
  year={2025}
}

@inproceedings{singh2024aya,
  title={Aya Dataset: An Open-Access Collection for Multilingual Instruction Tuning},
  author={Singh, Shivalika and Vargus, Freddie and D’souza, Daniel and Karlsson, B{\"o}rje and Mahendiran, Abinaya and Ko, Wei-Yin and Shandilya, Herumb and Patel, Jay and Mataciunas, Deividas and O’Mahony, Laura and others},
  booktitle={ACL},
  year={2024}
}

@article{lambert2024tulu,
  title={Tulu 3: Pushing frontiers in open language model post-training},
  author={Lambert, Nathan and Morrison, Jacob and Pyatkin, Valentina and Huang, Shengyi and Ivison, Hamish and Brahman, Faeze and Miranda, Lester James V and Liu, Alisa and Dziri, Nouha and Lyu, Shane and others},
  journal={arXiv preprint},
  year={2024}
}

@inproceedings{ethayarajh2024model,
  title={Model Alignment as Prospect Theoretic Optimization},
  author={Ethayarajh, Kawin and Xu, Winnie and Muennighoff, Niklas and Jurafsky, Dan and Kiela, Douwe},
  booktitle={ICML},
  year={2024}
}

@inproceedings{zeng2024token,
  title={Token-level direct preference optimization},
  author={Zeng, Yongcheng and Liu, Guoqing and Ma, Weiyu and Yang, Ning and Zhang, Haifeng and Wang, Jun},
  booktitle={ICML},
  year={2024}
}

@inproceedings{rafailov2023direct,
  title={Direct preference optimization: Your language model is secretly a reward model},
  author={Rafailov, Rafael and Sharma, Archit and Mitchell, Eric and Manning, Christopher D and Ermon, Stefano and Finn, Chelsea},
  booktitle={NeurIPS},
  year={2023}
}

% \newpage
\appendix
\section{Additional Experiment Results} \label{app:a}

\subsection{Hyperparameter Search} \label{app:a.1}
In SFTMix, we use the hyperparameter $\mu$ to control the regularization effect in the training loss $\ell_\text{SFTMix}$ and $\alpha$ to control the sampling distribution of $\lambda$.
To explore their impacts, we experiment with $\mu \in \{0.05, 0.1, 0.2, 0.5, 1\}$ and $\alpha \in \{0.1, 0.2, 0.5, 0.7, 1\}$ by instruction-tuning Llama on Alpaca-52K and UltraChat-200K. 
As shown in Table~\ref{tab:mu}, $\mu=0.2$ achieves the highest performance across most metrics. 
Similarly, Table~\ref{tab:alpha} shows that $\alpha=0.5$ yields the best overall score on MT-Bench and the highest LC WR on AlpacaEval-2.
Hence, we set $\mu = 0.2$ and $\alpha=0.5$ in \S\ref{sec:4} and \S\ref{sec:5}.
We also observe that performance remains stable for $\mu$ between $0.1$ and $0.2$ and $\alpha$ between $0.2$ and $0.5$.
However, performance declines when $\alpha$ approaches $0.1$ or $1$, where the Beta distribution becomes too flat and the interpolated examples deviate excessively from either endpoint.

In SFTMix, we apply Mixup regularization by interpolating between the most and least confident halves of the dataset. 
We adopt an equal split for its simplicity and ease of implementation, avoiding additional hyperparameters. 
To examine this design choice, we also fine-tune Llama-3.1-8B using Mixup between the most and least confident thirds. 
As shown in Table~\ref{tab:partition}, the equal split continues to yield the largest improvement over NTP. 
These results suggest that equal partitioning serves as a strong default, while more adaptive strategies remain a promising direction for future work.

\subsection{Human Evaluation on AlpacaEval-2} \label{app:a.2}
To complement the LLM-as-a-judge evaluation in \S\ref{sec:4.1}, we conduct a human evaluation following the setup in~\citep{zhao2024long}. 
Specifically, we use the $80$ instructions from the Vicuna subset in AlpacaEval-2 and compare responses generated by Llama, instruction-tuned on Alpaca-52K using either NTP or SFTMix, in a head-to-head fashion. 
As in~\citep{zhao2024long}, we instruct evaluators to disregard response length in their judgments.

We collect $200$ human preference judgments, where Llama instruction-tuned with SFTMix wins $42.5\%$ of the time, NTP wins $26.5\%$, and the remaining $31\%$ are ties.
This result aligns with our observation in \S\ref{sec:4.1} that SFTMix outperforms NTP in instruction-following SFT tasks.

\begin{table}[t!]
    \centering
    \setlength\tabcolsep{5pt}
\resizebox{\columnwidth}{!}{
    \begin{tabular}{c|ccc|cc}
        \toprule
            \multirow{2}{*}{$\boldsymbol{\mu}$} & \multicolumn{3}{c|}{\textbf{MT-Bench}} & \multicolumn{2}{c}{\textbf{AlpacaEval-2}} \\
            & \textbf{ST} & \textbf{MT} & \textbf{Overall} & \textbf{WR} & \textbf{LC WR} \\
        \midrule
        \midrule
            \multicolumn{6}{l}{\textbf{Dataset:} Alpaca-52K} \\
        \midrule
            $0.05$ & $5.0030$ & $3.9623$ & $4.4827$ & $4.7100$ & $9.9138$ \\
            $0.1$ & $5.0600$ & $4.0238$ & $4.5419$ & $4.7715$ & $10.0172$ \\
            $\underline{0.2}$ & $\textbf{5.2125}$ & $3.9525$ & $\textbf{4.5825}$ & $\textbf{4.9031}$ & $\textbf{10.3195}$ \\
            $0.5$ & $4.9606$ & $3.8968$ & $4.4287$ & $4.5092$ & $9.5034$ \\
            $1$ & $4.7050$ & $\textbf{4.1075}$ & $4.4062$ & $3.9450$ & $8.2856$ \\
        \midrule
            \multicolumn{6}{l}{\textbf{Dataset:} UltraChat-200K} \\
        \midrule
            $0.05$ & $6.1764$ & $4.9208$ & $5.5486$ & $5.0433$ & $8.5521$ \\
            $0.1$ & $\textbf{6.2988}$ & $5.2713$ & $5.7851$ & $5.0823$ & $9.1174$ \\
            $\underline{0.2}$ & $6.2750$ & $\textbf{5.3500}$ & $\textbf{5.8125}$ & $\textbf{5.1149}$ & $\textbf{9.3810}$ \\
            $0.5$ & $5.9701$ & $4.5899$ & $5.2800$ & $4.8991$ & $8.6712$ \\
            $1$ & $5.6621$ & $4.2490$ & $4.9556$ & $4.5661$ & $8.2908$ \\
        \bottomrule
    \end{tabular}
}
    % \vspace{-2.5mm}
    \caption{Hyperparameter search for $\mu$. We set $\mu = \underline{0.2}$ as the default in SFTMix, as it achieves the highest performance across most metrics.}
    % \vspace{-2mm}
    \label{tab:mu}
\end{table}

\begin{table}[t!]
    \centering
    \setlength\tabcolsep{5pt}
\resizebox{\columnwidth}{!}{
    \begin{tabular}{c|ccc|cc}
        \toprule
            \multirow{2}{*}{$\boldsymbol{\alpha}$} & \multicolumn{3}{c|}{\textbf{MT-Bench}} & \multicolumn{2}{c}{\textbf{AlpacaEval-2}} \\
            & \textbf{ST} & \textbf{MT} & \textbf{Overall} & \textbf{WR} & \textbf{LC WR} \\
        \midrule
        \midrule
            \multicolumn{6}{l}{\textbf{Dataset:} Alpaca-52K} \\
        \midrule
            $0.1$ & $5.1632$ & $3.8991$ & $4.5312$ & $4.8813$ & $10.0259$ \\
            $0.2$ & $\textbf{5.2237}$ & $3.9103$ & $4.5670$ & $\textbf{4.9112}$ & $10.1865$ \\
            $\underline{0.5}$ & $5.2125$ & $\textbf{3.9525}$ & $\textbf{4.5825}$ & $4.9031$ & $\textbf{10.3195}$ \\
            $0.7$ & $5.1209$ & $3.9291$ & $4.5250$ & $4.7912$ & $9.8247$ \\
            $1$ & $5.0826$ & $3.9200$ & $4.5013$ & $4.7526$ & $9.9835$ \\
        \midrule
            \multicolumn{6}{l}{\textbf{Dataset:} UltraChat-200K} \\
        \midrule
            $0.1$ & $5.7461$ & $4.9800$ & $5.3631$ & $4.9001$ & $8.5527$ \\
            $0.2$ & $6.2001$ & $\textbf{5.4137}$ & $5.8069$ & $5.0221$ & $9.1846$ \\
            $\underline{0.5}$ & $\textbf{6.2750}$ & $5.3500$ & $\textbf{5.8125}$ & $\textbf{5.1149}$ & $\textbf{9.3810}$ \\
            $0.7$ & $6.0138$ & $5.0012$ & $5.5075$ & $5.0811$ & $8.9312$ \\
            $1$ & $5.9112$ & $4.8641$ & $5.3877$ & $4.8813$ & $8.7512$ \\
        \bottomrule
    \end{tabular}
}
    % \vspace{-2.5mm}
    \caption{Hyperparameter search for $\alpha$. We set $\mu = \underline{0.5}$ as the default in SFTMix, as it achieves the highest performance across most metrics.}
    % \vspace{-2mm}
    \label{tab:alpha}
\end{table}

\begin{figure}[t!]
    \centering
    \includegraphics[width=\columnwidth]{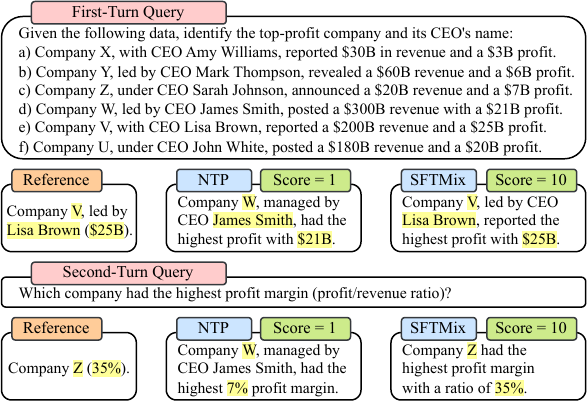}
    % \vspace{-6.5mm}
    \caption{A qualitative example from the extraction category in MT-Bench. Compared to its NTP-tuned counterpart, Llama instruction-tuned by SFTMix accurately interprets the queries from both turns and correctly extracts the relevant information from the prompt.}
    % \vspace{-2mm}
    \label{fig:qual}
\end{figure}

\begin{figure}[t!]
    \centering
    \includegraphics[width=\columnwidth]{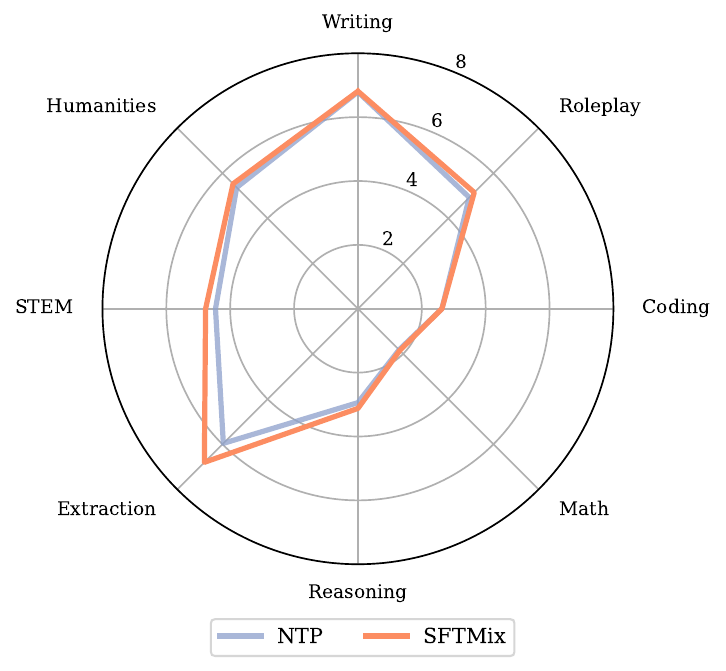}
    % \vspace{-6.5mm}
    \caption{Per-category MT-Bench scores for Llama-3.1-8B tuned on Alpaca-52K. SFTMix consistently outperforms NTP, notably in extraction, STEM, and roleplay.}
    % \vspace{-2mm}
    \label{fig:mt_cat}
\end{figure}

% \vspace{-0.5mm}
\subsection{MT-Bench Result Analysis} \label{app:a.3}
% \vspace{-0.25mm}

Figure~\ref{fig:qual} presents a test example from the extraction category in MT-Bench, showing responses generated by Llama-3.1-8B instruction-tuned on Alpaca-52K using either NTP or SFTMix. 
In this example, the LLM trained with SFTMix accurately interprets the instructions from both the first- and second-turn queries, correctly extracting the relevant information from the prompt. 
Notably, it succeeds in answering the second-turn query, which involves calculating the profit margin and performing a ratio comparison. 
In contrast, the LLM trained with NTP struggles to differentiate between revenue and profit, leading to incorrect responses in both turns.

To provide finer-grained insights, we report per-category MT-Bench scores for Llama-3.1-8B instruction-tuned on Alpaca-52K. 
As shown in Figure~\ref{fig:mt_cat}, SFTMix consistently improves over NTP across all categories, with the largest gains in extraction, STEM, and roleplay. 
However, both methods remain limited in coding, math, and reasoning, likely due to the lack of such data in Alpaca-52K.

\begin{table}[t!]
    \centering
    \setlength\tabcolsep{3pt}
\resizebox{\columnwidth}{!}{
    \begin{tabular}{cc|ccc|cc}
        \toprule
            \multirow{2}{*}{\textbf{Recipe}} & \multirow{2}{*}{\textbf{Mixup}} & \multicolumn{3}{c|}{\textbf{MT-Bench}} & \multicolumn{2}{c}{\textbf{AlpacaEval-2}} \\
            & & \textbf{ST} & \textbf{MT} & \textbf{Overall} & \textbf{WR} & \textbf{LC WR} \\
        \midrule
        \midrule
            NTP & - & $4.9100$ & $3.8150$ & $4.3625$ & $4.0714$ & $8.6528$ \\
            SFTMix & \underline{Half} & $\textbf{5.2125}$ & $3.9525$ & $\textbf{4.5825}$ & $\textbf{4.9031}$ & $\textbf{10.3195}$ \\
            SFTMix & Third & $5.1450$ & $\textbf{3.9850}$ & $4.5650$ & $4.2073$ & $8.8978$ \\
        \bottomrule
    \end{tabular}
}
    % \vspace{-2.5mm}
    \caption{Evaluation of the Mixup ratio in SFTMix. By default, SFTMix applies Mixup between the most and least confident \underline{halves}, yielding greater improvement.}
    % \vspace{-2mm}
    \label{tab:partition}
\end{table}

\begin{table}[t!]
    \centering
\resizebox{\columnwidth}{!}{
    \begin{tabular}{cc|ccc}
        \toprule
            \textbf{Language} & \textbf{Recipe} & \textbf{Win} & \textbf{Tie} & \textbf{Loss} \\
        \midrule
        \midrule
            \multirow{2}{*}{Chinese} & NTP & $30.4\%$ & $35.6\%$ & $34.0\%$ \\
            & SFTMix & $\textbf{38.8}\%$ & $33.2\%$ & $28.0\%$ \\
        \midrule
            \multirow{2}{*}{French} & NTP & $34.5\%$ & $24.5\%$ & $41.0\%$ \\
            & SFTMix & $\textbf{39.0}\%$ & $22.5\%$ & $38.5\%$ \\
        \bottomrule
    \end{tabular}
}
    % \vspace{-2.5mm}
    \caption{Multilingual evaluation of SFTMix. The method consistently improves win rates in both Chinese and French instruction tuning.}
    % \vspace{-2mm}
    \label{tab:multilingual}
\end{table}

% \vspace{-0.5mm}
\subsection{Multilingual Instruction Tuning} \label{app:a.4}
% \vspace{-0.25mm}

To assess SFTMix in multilingual settings, we fine-tune Llama-3.1-8B on the Chinese~\citep{peng2023instruction} and French~\citep{chen2023multilingualsift} variants of Alpaca-52K, following the setup in \S\ref{sec:4.1}. 
We then compare the resulting models with human-annotated baselines from Aya~\citep{singh2024aya} in the respective languages, using GPT-4o~\citep{hurst2024gpt} as the judge. 
As shown in Table~\ref{tab:multilingual}, SFTMix increases the win rate in both languages, demonstrating its cross-lingual generalizability.

\end{document}